\documentclass{article}

\usepackage[final]{neurips_2025}

\usepackage[utf8]{inputenc} 
\usepackage[T1]{fontenc}    
\usepackage{hyperref}       
\usepackage{url}            
\usepackage{booktabs}       
\usepackage{amsfonts}       
\usepackage{nicefrac}       
\usepackage{microtype}      
\usepackage{xcolor}         

\usepackage{fixltx2e}
\usepackage{multirow}
\usepackage{latexsym}
\usepackage{amssymb}
\usepackage{amsmath}
\usepackage{amsthm}
\usepackage{booktabs}
\usepackage{enumitem}
\usepackage{graphicx}
\usepackage{caption}
\usepackage{color}
\usepackage{cleveref}
\usepackage{diagbox}
\usepackage{bm}
\usepackage{algorithm}
\usepackage{algpseudocode}
\usepackage{makecell}
\newcounter{example}[section]
\newenvironment{example}[1][]
{
    \refstepcounter{example}
    \par
    \medskip
    \noindent 
    \textbf{Example~\theexample. #1} 
    \rmfamily
}
{\medskip}

\makeatletter
\newcommand{\myfnsymbol}[1]{%
  \expandafter\@myfnsymbol\csname c@#1\endcsname
}
\newcommand{\@myfnsymbol}[1]{%
  \ifcase #1
  \or 1
  \or 2
  \or 3
  \fi
}
\newcommand{\corresponding}{\@myfnsymbol{1}}
\makeatother

\title{Time Reversal Symmetry for Efficient Robotic Manipulations in Deep Reinforcement Learning}

\author{%
  Yunpeng Jiang\\
  \texttt{jyp9961@sjtu.edu.cn}\\
  Global College\\
  Shanghai Jiao Tong University\\
  \And
  Jianshu Hu\\
  \texttt{hjs1998@sjtu.edu.cn}\\
  Global College\\
  Shanghai Jiao Tong University\\
  \AND
  Paul Weng\thanks{Corresponding authors.}\\
  \texttt{paul.weng@dukekunshan.edu.cn}\\
  Digital Innovation Research Center\\
  Duke Kunshan University\\
  \And
  Yutong Ban$^*$ \\
  \texttt{yban@sjtu.edu.cn}\\
  Global College\\
  Shanghai Jiao Tong University\\
}

\newcommand{\methodname}{{TR-DRL}} 

\usepackage[textsize=tiny]{todonotes}

\newtoggle{final}
\toggletrue{final}
\newcommand{\pw}[1]{\iftoggle{final}{#1}{{\color{blue} #1}}}

\newtoggle{rebuttal}
\togglefalse{rebuttal} 
\newcommand{\jypmodify}[1]{\iftoggle{rebuttal}{{\color{brown} #1}}{#1}}

\newcommand{\St}{{\mathcal S}} 
\newcommand{\Ac}{{\mathcal A}} 
\newcommand{\T}{{T}} 
\newcommand{\Reward}{{R}} 
\newcommand{\Potential}{{\Phi}}

\newcommand{\J}{J} 

\newcommand{\Q}{{Q}} 

\newcommand{\eExpect}{{\hat{\mathbb E}}}

\algnewcommand{\IIf}[1]{\State\algorithmicif\ #1\ \algorithmicthen}
\algnewcommand{\EndIIf}{\unskip\ \algorithmicend\ \algorithmicif}

\makeatletter
\DeclareRobustCommand{\cev}[1]{%
  {\mathpalette\do@cev{#1}}%
}
\newcommand{\do@cev}[2]{%
  \vbox{\offinterlineskip
    \sbox\z@{$\m@th#1 x$}%
    \ialign{##\cr
      \hidewidth\reflectbox{$\m@th#1\vec{}\mkern4mu$}\hidewidth\cr
      \noalign{\kern-\ht\z@}
      $\m@th#1#2$\cr
    }%
  }%
}
\makeatother

\begin{document}

\maketitle

\footnotetext[1]{Project Page: \url{https://jyp9961.github.io/TR-DRL_project_page/}\label{footnote: project page}}
\footnotetext[2]{Source Code: \url{https://github.com/jyp9961/TR-DRL}\label{footnote: source code}}

\begin{abstract}

Symmetry is pervasive in robotics and has been widely exploited to improve sample efficiency in deep reinforcement learning (DRL).
However, existing approaches primarily focus on spatial symmetries—such as reflection, rotation, and translation—while largely neglecting temporal symmetries.
To address this gap, we explore time reversal symmetry, a form of temporal symmetry commonly found in robotics tasks such as door opening and closing.
We propose Time Reversal symmetry enhanced Deep Reinforcement Learning (\methodname), a framework that combines trajectory reversal augmentation and time reversal guided reward shaping to efficiently solve temporally symmetric tasks.
Our method generates reversed transitions from fully reversible transitions, identified by a proposed dynamics-consistent filter, to augment the training data.
For partially reversible transitions, we apply reward shaping to guide learning, according to successful trajectories from the reversed task.
Extensive experiments on the Robosuite and MetaWorld benchmarks demonstrate that {\methodname} is effective in both single-task and multi-task settings, achieving higher sample efficiency and stronger final performance compared to baseline methods.
Our project website and source code can be found in \ref{footnote: project page} and \ref{footnote: source code}.
\end{abstract}

\section{Introduction}
Deep reinforcement learning (DRL) is a powerful machine learning framework capable of solving complex tasks, with applications across robotics, quantitative trading, and video games.
Despite its successes, DRL often suffers from low sample efficiency and poor agent robustness.
To address these challenges, symmetry, a common property in many real-world scenarios, has been leveraged to improve both sample efficiency and agent performance.
Symmetry can be used to augment trajectories collected during training in both state-based \citep{Lin_2020, 10.1007} 
and image-based settings \citep{drqv2}.
\pw{Alternatively}, symmetry can be embedded directly into the network architecture, making it an inherent property of the model \citep{cohen2016groupequivariantconvolutionalnetworks, wang2022mathrmsoequivariant}.
\pw{In addition, it can be enforced as a regularization term \citep{hu2024revisitingdataaugmentationdeep, DrAC}.}

However, existing work \pw{(see Related Work in \Cref{sec:related})} predominantly focuses on spatial symmetries, 
such as translation, reflection, and rotation, while temporal symmetries, including time-reversal symmetry and time dilation, remain largely underexplored.
Intuitively, time-reversal symmetry corresponds to a reflection with respect to time, assuming actions can be reversed, which often holds in navigation tasks.
Time dilation occurs in certain robotics control problems when the agent can control the speed of action execution \citep{hu2024statenoveltyguidedactionpersistence}.
In this paper, we focus on leveraging time-reversal symmetry in robot manipulation tasks, where the agent controls the position and orientation of the end-effector.
Unlike spatial symmetries, where augmented samples typically remain valid, temporally reversed transitions may result in invalid transitions due to complex interactions between the robot and objects.

\begin{figure}[t]
    \centering
    \includegraphics[width=0.9\linewidth]{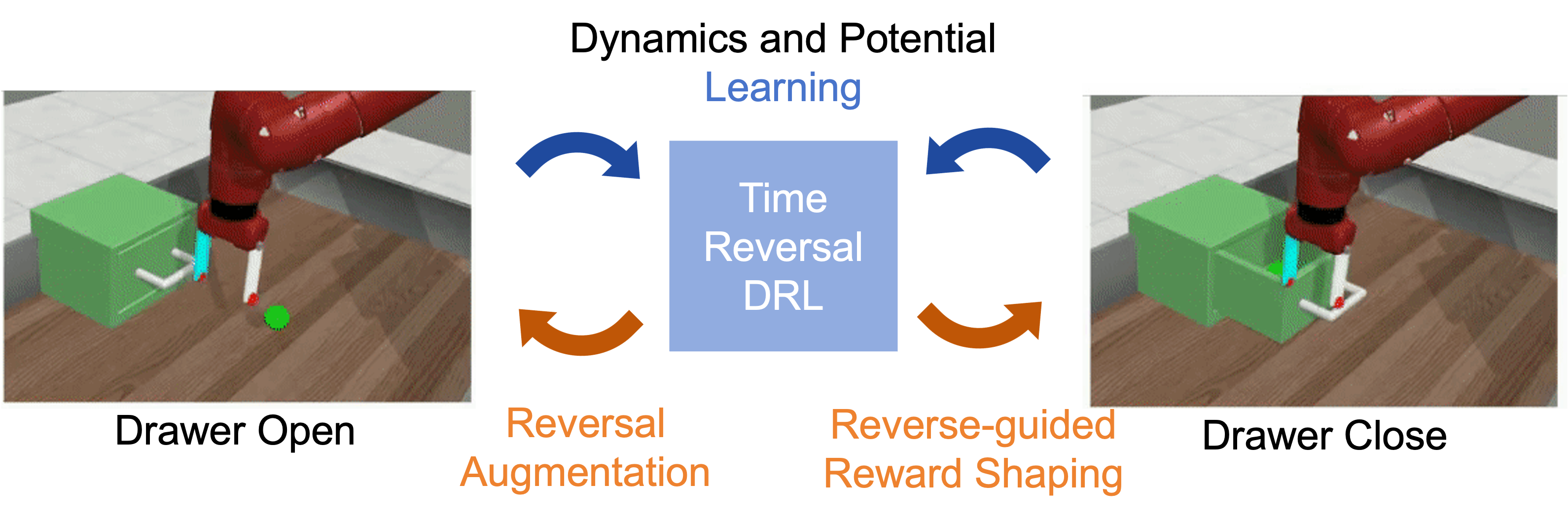}
    \caption{For a task pair, the proposed TR-DRL framework learns dynamics and potential models, leverages trajectory reversal augmentation with dynamics aware filtering and time reversal symmetry guided reward shaping, and boosts sample efficiency in both tasks.}
    \label{fig:teaser}
\end{figure}
\begin{figure}
    \centering
    \includegraphics[width=0.99\linewidth]{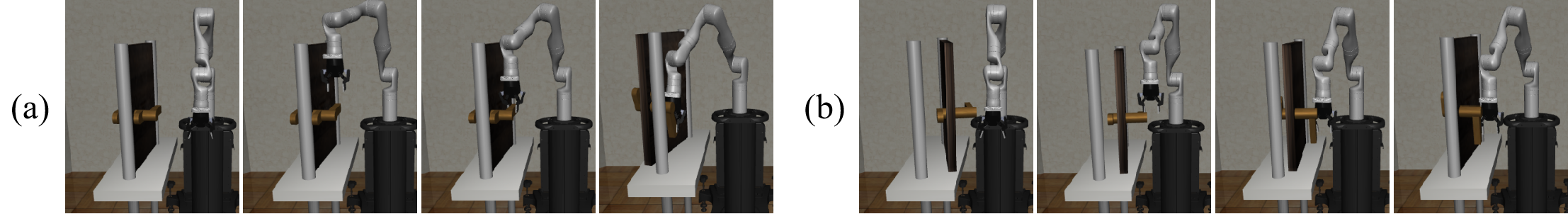}
    \caption{\jypmodify{Examples of fully and partially reversible trajectories.} (a) \jypmodify{Fully reversible:} An example of opening the door outward by grasping the handle; (b) \jypmodify{Partially reversible:} An example of closing the door from inward by pushing the door.}
    \label{fig:intro_door}
\end{figure}

Consider a task pair, door opening outward and door closing from outward.
An augmented trajectory of closing a door from outward (\Cref{fig:intro_door}(a) from right to left) can be generated by reversing a trajectory where the agent opens the door by grasping and moving the handle outward (\Cref{fig:intro_door}(a) from left to right).
In this case, the state pairs within the trajectory are fully reversible.
Then consider another task pair, door opening inward and door closing from inward.
When the agent closes the door by simply pushing it without grasping the handle (\Cref{fig:intro_door}(b)), reversing the trajectory becomes nontrivial.
This is because the agent cannot feasibly open the door without first grasping the handle, making the reversed transitions invalid.
\pw{However}, certain components of the state, such as the object state (e.g., the door’s opening angle), may still be reversible, even if the full transition is not.
Such \pw{cases correspond to} partial reversibility \pw{of} the transitions \jypmodify{where
the concept of state decomposition \citep{pitis2020counterfactualdataaugmentationusing} enables isolation of dynamically reversible components.}

To exploit \pw{(partial or full)} time reversal symmetry in DRL, we propose a general framework (see \Cref{fig:teaser}) that incorporates two complementary techniques\pw{, which can accelerate training for a pair of related tasks.}
For full time reversal symmetry, we learn an inverse dynamics model to obtain the reversed actions and generate the augmented transitions \pw{when training in both tasks.}
To ensure the validity of these reversed transitions, we additionally train a forward dynamics model to filter out transitions that violate the true system dynamics.
For \pw{partial time reversal symmetry}, the reversible component of the state can be intuitively used to guide policy learning.
We leverage this form of symmetry through reward shaping, encouraging the agent \pw{for one task} to follow trajectories that resemble the reversed versions of successful trajectories from the \pw{other} task.

\paragraph{Contributions}
Our contributions \pw{can be summarized as follows}:
\begin{enumerate}[label=(\roman*),wide]
    \item \pw{
Based on (full) time reversal symmetry (\Cref{sec:background}), we introduce
the novel notion of partial time reversal symmetry (\Cref{sec:problem settings}) to exploit temporal symmetry in more general settings (e.g., when objects are pushed).}

    \item \pw{
    We propose two techniques (\Cref{sec:trajectory reversal,sec:time reversal reward shaping}) to exploit time reversal symmetry:}
\begin{itemize}
    \item 
\pw{For full time reversal symmetry, t}ransitions identified as reversible by a \pw{trained} dynamics-aware filter are augmented to improve the sample efficiency of DRL algorithms.
    \item 
\pw{For partial time reversal symmetry, a} reward shaping mechanism 
\pw{exploits} transitions \pw{from successful trajectories} \pw{to guide the training of the DRL agent}.
\end{itemize}
    \item 
We conduct extensive experiments \pw{(\Cref{sec: experimental results})} on standard robotics benchmarks (Robosuite, Metaworld) demonstrating that our approach significantly improves both sample efficiency and final performance compared to baseline methods.
An ablation study further validates our design choices and highlights the contributions of each component within our framework.
\end{enumerate}

\section{Related Works} \label{sec:related}
We summarize related works from two research directions in DRL, which include symmetry, and reward shaping techniques. 
For symmetry, we divide it into spatial symmetry and time reversal symmetry.

\paragraph{Spatial Symmetry in DRL}
Spatial symmetry, including reflection, rotation, and translation, are extensively exploited in DRL. 
These symmetries enable the generation of synthetic transitions from a single environment interaction, effectively improving sample efficiency.
For example, prior works \citep{Lin_2020, corrado2024understandingdynamicsinvariantdataaugmentations, corrado2024guideddataaugmentationoffline} have shown that applying spatial augmentations such as reflection, rotation, and translation significantly boosts sample efficiency in state-based robotics control tasks.
In image-based RL, translation symmetry has also been widely adopted to enhance performance \citep{drqv2, DA_DRL, hu2024revisitingdataaugmentationdeep}.
\jypmodify{Data augmentation methods can also improve robustness to noise \citep{sinha2021s4rlsurprisinglysimpleselfsupervision, qiao2021efficientdifferentiablesimulationarticulated}.}
Moreover, spatial symmetry can be embedded directly into the neural network architecture through equivariance \citep{cohen2016groupequivariantconvolutionalnetworks, wang2022mathrmsoequivariant, wang2023surprisingeffectivenessequivariantmodels}, ensuring that the network respects these symmetries by design.
This architectural integration reduces training time and improves generalization across diverse inputs.
In contrast to these prior efforts, our work focuses on exploiting time-reversal symmetry, a form of temporal symmetry that remains underexplored in DRL.

\paragraph{Time Reversal Symmetry in DRL}
Time reversal symmetry has been leveraged for data augmentation 
\citep{barkley2023investigationtimereversalsymmetry} and for learning dynamics-consistent latent representations from images \citep{cheng2023lookbeneathsurfaceexploiting}.
In contrast to simply negating actions in reversed transitions \citep{barkley2023investigationtimereversalsymmetry, yao2023learningsymmetrymetareinforcementlearning}, our approach employs a more sophisticated strategy to derive reversed actions, making it applicable to a broader range of environments.
Moreover, our method focuses on state-based control tasks, where full state information is available, eliminating the need of learning latent representations from visual observations.

Some existing works focus on exploring reversibly from goal states, utilizing the time symmetry to enhance the agent's exploration towards desired states.
Starting from a goal state, the agent explores by imagining reversal steps \citep{edwards2018forwardbackwardreinforcementlearning} or predicting preceding states leading to goals \citep{goyal2019recalltracesbacktrackingmodels}.
Instead of using imagined trajectories, true trajectories starting from goal states are given in TRASS \citep{nair2020timereversalselfsupervision}, and the agent learns from the reversed trajectories.
Unlike these works, we leverage time reversal symmetry not only from goal states but for every transition in the trajectory, enabling a broader application of time symmetry across the entire state space.

Other prior works focus on enhancing the reversibility of the agent, exploring strategies to ensure that agents can backtrack or reset their actions to avoid irreversible states.
For instance, 
\citet{grinsztajn2021turningbackselfsupervisedapproach} propose to distinguish reversible from irreversible actions to improve decision-making in DRL.
This distinction enables agents to prioritize reversible actions that are safer, as they guarantee the ability to backtrack if needed.
Furthermore, \citet{eysenbach2017leavetracelearningreset} propose learning a reset policy alongside the normal policy to prevent agents from entering non-reversible states, ensuring safety in exploration phase and achieving better training efficiency.
While their reset policy sets initial state as the ending state of the current policy and the goal state as the task's starting point, our method treats two reversible tasks independently, with initial and goal states defined separately for each task.
Additionally, our method is orthogonal to theirs and can be integrated to enhance the training of their reset policy.

\paragraph{Reward Shaping in DRL}
Reward shaping is a powerful technique for enhancing the efficiency of DRL algorithms \citep{ibrahim2024comprehensiveoverviewrewardengineering}, as it guides agents toward desired behaviors.
The idea of using shaped rewards to guide learning naturally aligns with our objective of leveraging reversed trajectory in tasks of time reversal symmetry.
However, reward shaping has not yet been explored in the context of time reversal symmetry.
In this work, we exploit time reversal symmetry by training a potential function guided by reversed trajectories.
Potential-based reward shaping \citep{10.5555/645528.657613} involves defining a potential function over the state space, which captures the agent's desired progress toward the goal.
Importantly, the optimal policy remains unchanged with potential-based reward shaping, providing a theoretical foundation for its application in our method.

\section{Background}\label{sec:background}
In this section, we recall the framework of deep reinforcement learning (DRL), the soft actor-critic algorithm, the concept of time reversal symmetry\pw{,} and potential-based reward shaping \pw{in reinforcement learning (RL)}.


\paragraph{Deep Reinforcement Learning (DRL)}
For any set $\mathcal X$, $\Delta(\mathcal X)$ denotes the set of probability distributions over $\mathcal X$.
A Markov Decision Process (MDP) \pw{model} $M = (\St,\Ac, \Reward, \T, \rho_0, \gamma)$ is defined by a set of state $\St$, a set of action $\Ac$, 
a reward function $\Reward: \St \times \Ac \rightarrow \mathbb{R}$, 
a transition function $\T:\St \times \Ac \rightarrow \Delta(\St)$, a probability distribution over initial states $\rho_0 \in \Delta(\St)$, and a discount factor $\gamma \in \pw{[0, 1]}$.
In \pw{RL}, the agent learns a policy $\pi(\cdot \mid s) \in \Delta(\Ac)$ by interacting with the environment, aiming to maximize the expected return $\J = 
 \mathbb{E}_\pi[\sum_{t=0}^\infty \gamma^tr_t\mid s_0 \sim \rho_0]$, where $\mathbb{E}_\pi$ denotes the expectation over $\pi$ and $r_t$ is the reward that the agent obtains at each timestep $t$.

\paragraph{Soft Actor-Critic (SAC)}
Maximum entropy reinforcement learning (RL) addresses standard RL problems using an alternative objective that explicitly encourages stochastic policies.
The objective combines cumulative reward with an entropy term:
$\J = \eExpect_\pi[\sum_{t=0}^\infty \gamma^t r_t+\alpha H(\pi(\cdot \mid s_t))]$, where $\gamma$ is the discount factor, $\alpha$ is a trainable coefficient of the entropy term, and $H(\pi(\cdot \mid s_t))$ represents the entropy of of the policy distribution $\pi(\cdot \mid s_t)$. 
The Soft Actor-Critic (SAC) algorithm \citep{haarnoja2018softactorcriticoffpolicymaximum} optimizes this objective by training the actor $\pi_\theta$ and critic $\Q_\psi$ with the following losses:
\begin{equation}
\label{eq:SAC loss}
\begin{split}
    &L_\pi(\theta) = \eExpect_{s_t\sim \mathcal{D},a \sim \pi}[\alpha\log \pi_\theta(a \mid s_t)-Q_\psi(s_t,a)],\\
    &L_Q(\psi) = \eExpect_{s_t,a_t\sim \mathcal{D}}[(Q_\psi(s_t,a_t)-\hat{Q}(s_t,a_t))^2],
\end{split} 
\end{equation}
where $\hat{Q}(s_t,a_t)=r_t+\gamma Q_{\bar{\psi}}(s_{t+1},a_{t+1})-\alpha\log{\pi_{\theta}(a_{t+1}|s_{t+1})}$, which is the target Q-value computed using a target network, and $a_{t+1} \sim \pi_{\theta}(\cdot \mid s_{t+1})$. Here, $\theta$, $\psi$ and $\bar\psi$ represent the parameters of the actor, the critic and the target critic respectively, while $\mathcal D$ represents the replay buffer. 
To stabilize training, the weights of the target network are updated as an exponential moving average of the online critic network’s weights.

\paragraph{Time Reversal Symmetry in DRL}
\pw{Given an involution\footnote{Recall an involution is a one-to-one mapping, which is its own inverse.} $f: \St\times\Ac\times\St \to \St\times\Ac\times\St$, a}n MDP \pw{satisfies (full) time reversal symmetry (adapted from \citet{barkley2023investigationtimereversalsymmetry})}
if for all $s_t, s_{t+1} \in \St$, 
\begin{equation}
\label{eq:definition fully reversible}
    \pw{\T}(s_{t+1} \mid s_{t}, a_t) = \pw{\T}(\cev{s}_{t} \mid \cev{s}_{t+1}, \cev{a}_t).
\end{equation}
\pw{where} $(\cev{s}_{t+1}, \cev{a}_t, \cev{s}_t) = f(s_t, a_t, s_{t+1})$ and \jypmodify{$\cev{\cdot}$ denotes time reversal operation on $\mathcal{S}$ or $\mathcal{A}$.}
Intuitively, involution $f$ represents the symmetry that reverses the passage of time.
Note that in some situations, it can be simply written as $f(s, a, s') = (f_\St(s), f_\Ac(a), f_\St(s'))$ using an involution $f_\St$ over states and an involution $f_\Ac$ over actions.
An example of time reversal symmetry in physical system is the transformation of position $p$, momentum $q$, and the applied force $a$.
The involution $\pw{f_\St}$ transforms state $s = (q, p)$ \pw{into} $\pw{f_\St}(s) = (q, -p)$, preserving position while negating momentum\jypmodify{, which is a common phenomenon in physical systems}.
For the action $a$, $\pw{f_\Ac}$ reverses the applied force such that $\pw{f_\Ac}(a) = -a$.
This ensures that the dynamics remain consistent under time reversal.

\paragraph{Potential-Based Reward Shaping}
We recall the concept of potential-based reward shaping proposed by \citet{10.5555/645528.657613}.
A shaping reward function $\mathcal{F}: \St \times \Ac \times \St \rightarrow \mathbb{R}$ is potential-based if there exists a real-valued function $\Potential: \St \rightarrow \mathbb{R}$ such that for all $s \in \St$, $a \in \Ac$, $s^\prime \in \St$,
\begin{equation}
    \mathcal{F}(s, a, s^\prime) = \gamma \Potential(s^\prime) - \Potential(s),
    \label{eq:potential definition}
\end{equation}
This condition is necessary and sufficient to ensure \pw{that an} optimal policy of the modified MDP $M^\prime=(\St, \Ac, \Reward+\mathcal{F}, \T, \rho_0, \gamma)$ 
\pw{remains optimal in}
the original MDP $M=(\St, \Ac, \Reward, \T, \rho_0, \gamma)$.

\section{Methodology}
\label{sec:method}
In this section, we \pw{first} define the problem \pw{set-up considered in this paper and} introduce two types of time-reversal symmetry, full \pw{or} partial time reversal \pw{symmetries}, \pw{for which we} provide illustrative examples in robotics (\Cref{sec:problem settings}).
\begin{figure}[t]
    \centering
    \includegraphics[width=0.99\linewidth]{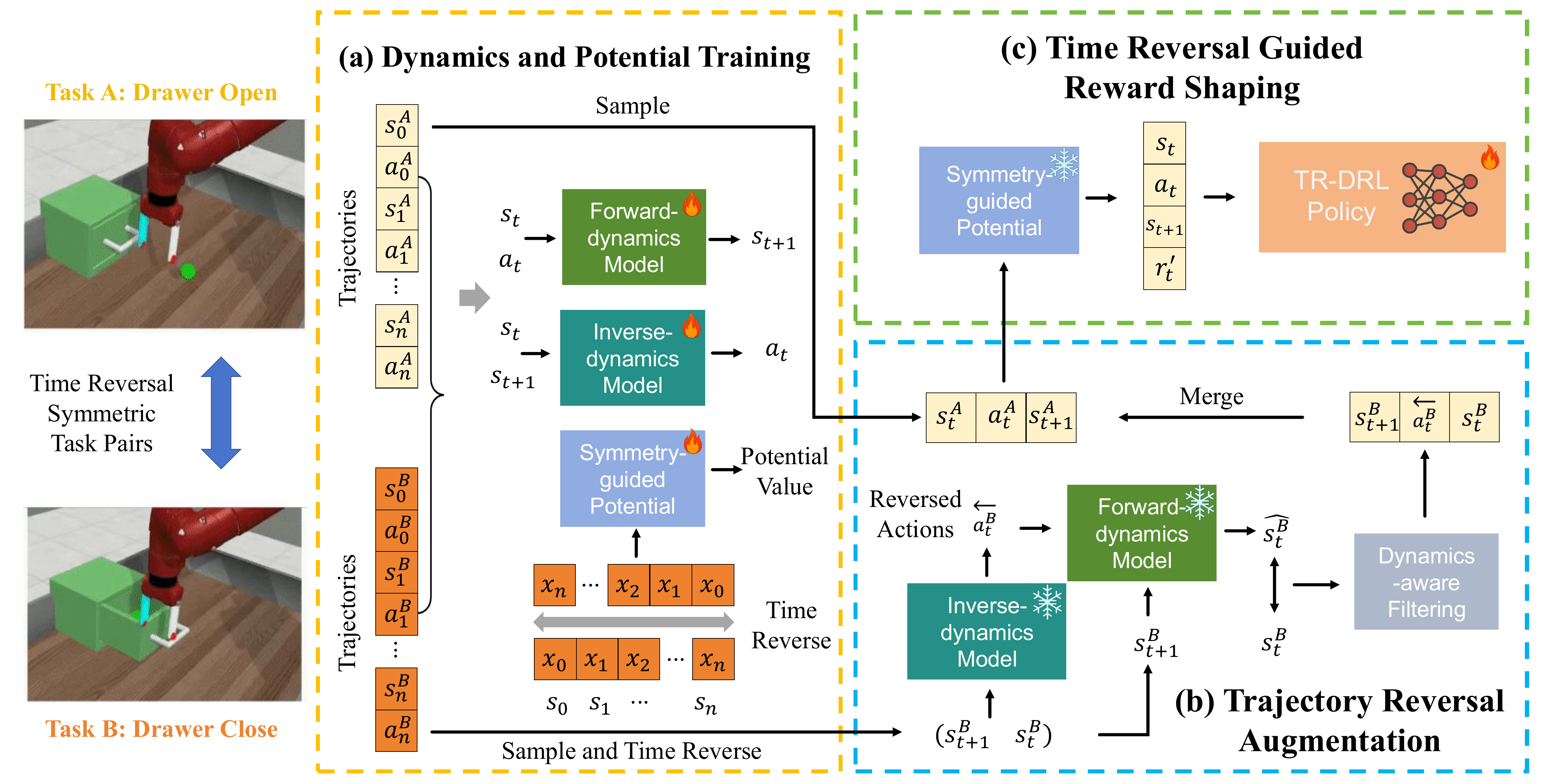}
    \caption{\textbf{Overview of our \methodname}. 
    We learn dynamics and potential models, apply reversal augmentation on transitions from the reversed task, and apply time reversal symmetry guided reward shaping on all transitions.}
    \label{fig:overview}
\end{figure}
We then propose a generic method, as shown in \Cref{fig:overview}, which applies trajectory reversal augmentation (\Cref{sec:trajectory reversal}) on \pw{fully reversible} transitions identified by our proposed dynamics-consistent filter, and employs reward shaping (\Cref{sec:time reversal reward shaping}) guided by \pw{partially reversible} transitions.


\subsection{Problem Formulation}
\label{sec:problem settings}

\pw{In this paper, we assume that the RL agent aims at learning to solve (at least) two related tasks in the same environment (e.g., door opening/closing or peg insertion/removal).
For such a pair of tasks, the RL agent may learn in a more data-efficient way by exploiting full time reversal (FTR) symmetry and partial time reversal (PTR) symmetry (see definition below).
While \citet{barkley2023investigationtimereversalsymmetry} assume that the FTR symmetry holds globally and that the involution to transform actions is known, which makes this temporal symmetry property too restrictive and difficult to apply in practice, we do not make these two assumptions, which allows us to consider scenarios like the next example.
However, the challenge now is to detect when \Cref{eq:definition fully reversible} holds and learn to recover reverse actions $\cev{a}_t$.}

\begin{example}
\label{example:full time reversal symmetry}
\pw{Consider} a pair of manipulation tasks: peg insertion and peg removal. 
\pw{Assume}
end-effector position control and \pw{that} the state includes the positions of both the end-effector and the \pw{object (i.e., peg)}.
When the robot arm holds and moves the peg towards the hole, for a transition $(s_t, a_t, s_{t+1})$, reversing the action enables the agent to move from $\pw{s_{t+1}}$ back to $\pw{s_{t}}$ without violating the true dynamics.
This means that all transitions along this trajectory exhibit FTR symmetry.
\pw{However, contacts and frictions prevent the definition of involution $f$ and if the peg can be dropped, the transitions are naturally not reversible anymore.}
\end{example}

\pw{While the relaxation of these two assumptions extend the applicability of FTR symmetry, for many pairs of tasks, an even weaker notion of temporal symmetry may be needed.
We therefore introduce the novel notion of partial time reversal (PTR) symmetry:}

\paragraph{Partial Time Reversal (PTR) Symmetry}
\pw{Assume that a state $s \in \St$ (resp. $s' \in \St$) can be decomposed into two parts $(x, y) \in \mathcal X \times \mathcal Y$ (resp. $(x', y') \in \mathcal X \times \mathcal Y$) and that an involution $f_{\mathcal X} : \mathcal X \to \mathcal X$ is given.
}
\pw{A} pair of states $(s, s') \in \pw{\St^2}$ \pw{satisfies PTR symmetry} if there \pw{exist} $(\cev{y}, \cev{y}') \in \pw{\mathcal{Y}^2}$ \pw{and} $(a, \cev{a}) \in \pw{\Ac^2}$ such that:
\begin{equation}
\label{eq:definition partially reversible}
\T(s' \mid s, a) = \T(\cev{s} \mid \cev{s}', \cev{a}),
\end{equation}
\pw{where $\cev{x} = f_{\mathcal X}(x)$, 
$\cev{x}' = f_{\mathcal X}(x')$, 
$\cev{s} = (\cev{x}, \cev{y})$, and 
$\cev{s}' = (\cev{x}', \cev{y}')$.
Intuitively, $\mathcal X$ is the part that is reversible (e.g., containing object state information).
Using this weaker property, we can now account for scenarios like the following example:}


\begin{example}
\label{example:partial time reversal symmetry}
\pw{Consider} another pair of tasks: door opening and door closing inward, with a similar definition of state and action  \pw{spaces as in \Cref{example:full time reversal symmetry}}.
In the door closing task, the agent learns to close the door by pushing it, without grasping the handle.
Along this trajectory, the transitions \pw{do not satisfy FTR symmetry}, as there does not exist an action that allows the robot arm to pull the door without grasping the handle.
However, the object \pw{(i.e. door)} state remains reversible.
We can find corresponding state pairs with reversed object state in the trajectories of door opening tasks.
These pairs reflect PTR symmetry, as only the object component of the state is reversible.   
\end{example}

\pw{In the next two subsections, we explain how to exploit FTR and PTR symmetries in DRL.}

\subsection{Trajectory Reversal Augmentation with Dynamics-Aware Filtering}
\label{sec:trajectory reversal}

In this section, we introduce how we augment the fully reversible transitions and how these fully reversible transitions are detected by a dynamics-consistent filter.
Given a pair of tasks with time reversal symmetry, 
any transition $(s, a, s')$ exhibiting FTR symmetry defined above can be augmented by generating its reversed transition $(s', \cev{a}, s)$ and incorporating it into DRL training.
Now the problem to be solved is finding $\cev{a}$.
In some robotics tasks, a straightforward choice of $\cev{a}$ is to negate the action which corresponds to reversing forces or torques, i.e. $\cev{a} = -a$ \citep{barkley2023investigationtimereversalsymmetry}.
However, it does not work for tasks involving contact dynamics or non-linear effects.
To address \pw{this}, we propose a more general approach by learning an inverse dynamics model $h$, \pw{represented} by a neural network:
\begin{equation}
    a = h(s,s'),
\end{equation}
which \pw{is} trained \pw{using transitions collected during RL training by minimizing the following loss:}
\begin{equation}
\label{eq:inv loss}
    L_{h} = \eExpect_{(s,a, s')\sim \mathcal{D}}[(h(s, s') - a)^2],
\end{equation}
where $\eExpect$ is an empirical mean estimating the expectation over \pw{the} true data distribution \pw{and} $\mathcal{D}$ denotes the replay buffer \pw{containing transitions} $(s, a, s')$.
Since the pair of reversible tasks share the same underlying dynamics, a single inverse dynamics model can be trained jointly using transitions from both tasks.
This shared model ensures the accurate inverse predictions when applied to the reversed task.

Note that trajectory reversal augmentation can only be applied directly on fully reversible transitions.
To identify such transitions, an additional dynamics-consistent filter is introduced to select appropriate samples from the replay buffer.
This filtering is achieved by training a forward dynamics model $g$ on transitions from both tasks \pw{by minimizing the following loss:}
\begin{equation}
\label{eq:for loss}
    L_{g} = \eExpect_{(s,a, s')\sim \mathcal{D}}[(g(s, a) - s')^2].
\end{equation}
This model allows us to verify whether a reversed transition $(s', \cev{a}, s)$ is consistent with the underlying dynamics.
In particular, for a reversed transition $(s', \cev{a}, s)$, we feed the state $s'$ and action $\cev{a}$ into the forward dynamics model \pw{$g$} to get the predicted state $\hat{s}$:
\begin{equation}
    \hat{s} = g(s', \cev{a}) = g(s', h(s',s)).
\end{equation}
The error between the predicted state $\hat{s}$ and the true state $s$ serves as a measure of feasibility.
Only when this prediction error $\mid\mid s - \hat{s} \mid\mid$ is below a predefined threshold $\beta$, the reversed transition is considered valid and included in \pw{the} training for the reversed task.

\subsection{Time Reversal Symmetry Guided Reward Shaping}
\label{sec:time reversal reward shaping}
In scenarios where \pw{not all} transitions \pw{are fully reversible}, trajectory reversal augmentation may become \pw{less} effective.
\pw{As an illustration, consider} the task mentioned in \Cref{example:partial time reversal symmetry}.
In such cases, most reversed transitions are filtered out by the dynamics-consistent filter, since the agent cannot reverse the action (i.e., from "push the door" to "pull the door") without first grasping the handle.
However, we can still exploit partial time reversal symmetry to improve sample efficiency.
In many tasks, the object state, 
such as the position of a door or the placement of a peg, remains reversible, while irreversibility arises primarily from the agent state, such as joint angles or gripper force.

To exploit this separation, we examine the relationship between object states in the trajectories of a pair of partially reversible tasks.
Let $x_t$ denote the object-related component of the full state $s_t$ at time step $t$.
Given a high-reward trajectory $\tau = (s_{0}, s_{1}, ..., s_{n})$ from one task, 
another trajectory $\cev{\tau} = (\cev{s}_{0}, \cev{s}_{1}, ..., \cev{s}_{n})$ from the reversed task should likewise receive a high reward if it achieves the reversed sequence $(x_{n}, x_{n-1}, ..., x_{0})$, 
or a partial reward if it accomplishes only a portion of the reversed sequence.
This observation raises a key question: can we leverage the partially reversible time symmetry, such that the reversible object-related components can be used to accelerate agent training?

\pw{As an answer to this question, we}
propose time reversal symmetry guided reward shaping.
Here we employ potential-based reward shaping \citep{10.5555/645528.657613} \pw{since it preserves} policy optimality and directly operates on \pw{states}.
To fully utilize multiple successful trajectories, we propose to train a potential model $\Potential$ for the reversed task, which maps the object state $x_{t}$ to a potential value $\Potential(x_{t})$.
As discussed, for the reversed trajectory containing sequences from $x_{t}$ to $x_{0}$, the potential values of these object states should increase in the reversed task.
Therefore, the object states along this reversed trajectory are labeled with potential values ranging from $0$ to $1$, and used to train the reversed task.
Here, a linear function can be used to interpolate potential values between the start and end states.
The potential model $\Potential$, is then trained to minimize the following loss:
\begin{equation}
\label{eq:pot loss}
\begin{aligned}
    &L_{\Potential} = \eExpect_{\tau = (s_{0}, s_{1}, ..., s_{n})\sim \mathcal{B}}\left[\left(\Potential(x_{t}) - \frac{n-t}{n}\right)^2\right],\\
    &s_t = [x_t,y_t], t \in (0,...,n)
\end{aligned}
\end{equation}
where $\mathcal{B}$ denotes the dataset which includes high-reward trajectories.

Based on \Cref{eq:potential definition}, the reward for each transition $(\cev{s}_t, \cev{a}_t, \cev{s}_{t+1}, \cev{r}_t)$ in the reversed task is reshaped as $\cev{r}_t + \gamma \Potential(x_{t+1}) - \Potential(x_t)$ during training.
This potential-based reward shaping mechanism encourages the agent to align the trajectory of object states with the reversed successful trajectory by dynamically shaping rewards based on the potential values.
In the door tasks, for example, a closed-to-open trajectory, reversed from a successful open-to-closed trajectory, guides the door opening agent by training $\Potential$ to predict low potential values for closed door states and high potential values for open door states.
Since potential-based reward shaping assigns a distinct potential value to each object state within the trajectory, reflecting its proximity to the success state, this smooth progression of potential values improves agent training by offering step-by-step guidance towards the goal state.

\subsection{\methodname \ Algorithm}
\pw{Our proposed techniques to exploit time reversal symmetry can be integrated in various DRL algorithms.
For concreteness, an}
example with SAC is presented in \Cref{alg:pseudocode}.
The training process alternates between two reversible tasks.
First, \pw{the} agents collect data from their respective environments. 
Then transitions from both environments are used to train the forward and inverse dynamics models.
Meanwhile, successful trajectories are employed to update the potential models.
During agent training, we augment \pw{the} original samples from the agent's current task with reversed samples from the reversible task 
via trajectory reversal augmentation and dynamics-aware filtering.
Further, we apply time reversal symmetry guided reward shaping to reshape rewards of all the transitions.
Finally, we update the agent with DRL loss.
\begin{algorithm}[t]
\caption{\methodname}
\label{alg:pseudocode}
\textbf{Required}: a pair of reversible tasks $(A, B)$, total number of training episodes $N$, total number of timesteps in one episode $T$.
\begin{algorithmic}[1] 
\State Initialize empty replay buffers $\mathcal{D_A}$ and $\mathcal{D_B}$. Initialize actor $\pi$ and critic $Q$.
\State Initialize potential models. Initialize forward and inverse dynamics models.
\For{$n=0 \dots N$}
    \For{$t=0 \dots T$}
        \State Alternate the following training steps between $A$ and $B$. 
        \State // Task $A$:
        \State The agent interacts with the environment and save the transition in replay buffer $\mathcal D_A$.
        \State Update forward and inverse dynamics models using \Cref{eq:inv loss} and \Cref{eq:for loss}.
        \State Update potential models using \Cref{eq:pot loss}.
        \State Sample two minibatches $d_A$ and $d_B$ from $\mathcal{D_A}$ and $\mathcal{D_B}$.
        \State Generate $d_{B, aug}$ from $d_B$ by reversal augmentation with dynamics-aware filtering.
        \State Apply time reversal symmetry guided reward shaping on $d_A \cup d_{B, aug}$.
        \State Update the actor and critic, $\pi$ and $Q$, with $d_A \cup d_{B, aug}$ using \Cref{eq:SAC loss}.
        \State // Task $B$: ...
    \EndFor
\EndFor

\end{algorithmic}
\end{algorithm}

\section{Experimental Results}
\label{sec: experimental results}
To demonstrate the effectiveness of our proposed method, we conduct comprehensive experiments to assess the performance of our approach in both single-task and multi-task settings.
We also run ablation study on our method to illustrate the design choice of different components.

\paragraph{Experimental setup}
\label{sec:experimental setup}
To validate our method, we evaluate our method in 60 environments from two standard robotics control benchmarks, Meta-World \citep{yu2021metaworldbenchmarkevaluationmultitask} and Robosuite \citep{zhu2025robosuitemodularsimulationframework}.
Detail introductions and example figures of these environments are provided in \Cref{appendix: environments}.
We use SAC \citep{haarnoja2018softactorcriticoffpolicymaximum}, multi-task SAC \citep{yu2021metaworldbenchmarkevaluationmultitask}, and multi-headed SAC \citep{yu2021metaworldbenchmarkevaluationmultitask} as the baselines for comparison.
Hyperparameters, such as network architecture and learning rates, are listed in \Cref{appendix: hyperparameter}.
We use sparse rewards in all our experiments, which makes learning challenging for the agent.
To mitigate this, we initialize the agent’s replay buffer with 10 expert demonstration trajectories, providing guidance to agent's exploration.
Unless specified, all reported scores are averaged over five runs, with standard deviations included in the results.
Throughout each run, the agent is evaluated every 20 training episodes by calculating the average success rate of 20 evaluation episodes.
To present the aggregated performance, we compute the inter-quartile mean (IQM) as proposed by \citet{agarwal2021deep}.

\subsection{Main results}
\begin{figure}[th]
    \begin{minipage}{0.26\textwidth}
        \centering
        \includegraphics[width=0.95\linewidth]{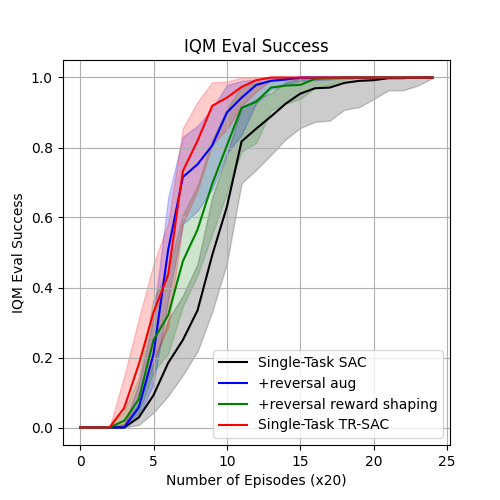}
    \end{minipage}
    \begin{minipage}{0.73\textwidth}
        \centering
        \small
        \begin{tabular}{c c c c c}
            \toprule
               \multirow{2}{*}{Method} & \multicolumn{4}{c}{Number of Environment Transitions} \\
                & 50k & 100k & 150k & 200k\\
            \midrule
              Single-Task SAC & \makecell{0.11$\pm$0.07} & \makecell{0.62$\pm$0.16} & \makecell{0.93$\pm$0.07} & \makecell{0.97$\pm$0.03} \\
            \midrule
              \makecell{+reversal aug} & \makecell{0.22$\pm$0.10} & \makecell{0.88$\pm$0.10} & \makecell{0.99$\pm$0.01} & \makecell{1.00$\pm$0.00} \\
            \midrule
              \makecell{+reversal\\reward shaping} & \makecell{0.26$\pm$0.10} & \makecell{0.79$\pm$0.12} & \makecell{0.97$\pm$0.03} & \makecell{1.00$\pm$0.00} \\
            \midrule
              \makecell{Single-Task\\ TR-SAC \textbf{(Ours)}} & \textbf{\makecell{0.33$\pm$0.14}} & \textbf{\makecell{0.92$\pm$0.07}} & \textbf{\makecell{1.00$\pm$0.00}} & \textbf{\makecell{1.00$\pm$0.00}} \\
            \bottomrule
        \end{tabular}
    \end{minipage}
    \hfill
    
    \begin{minipage}{0.99\textwidth}
        \includegraphics[width=0.99\linewidth]{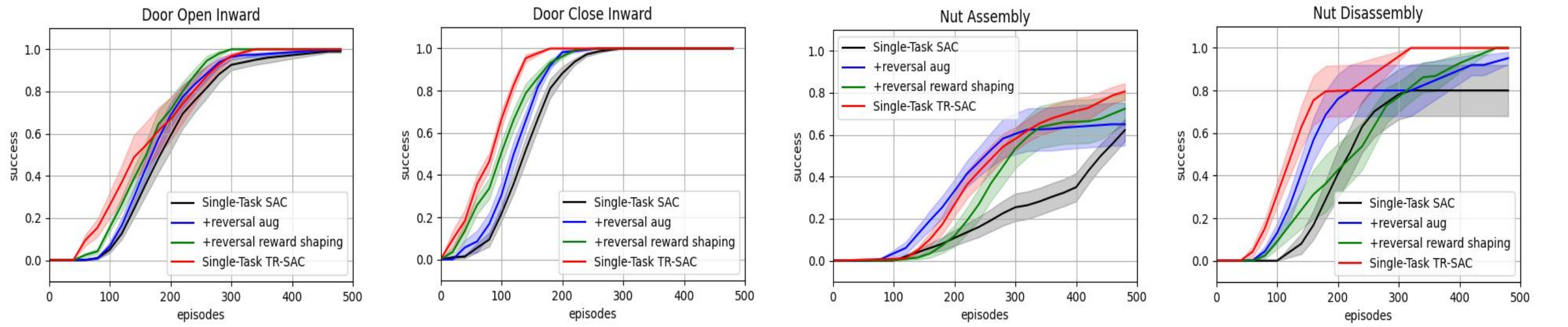}
    \end{minipage}
    \caption{
        \textbf{Results for single-task setting in 10 environments from Robosuite.}
        Top: Plots and table for IQM of success rate.
        Bottom: Curves of success rate in two pair of reversible tasks.
        "Single-Task SAC": baseline; "+reversal aug": trajectory reversal augmentation with dynamics-aware filtering; "+reversal reward shaping": time reversal symmetry guided reward shaping.}
    \label{fig:IQM_rev_aug_filter_rev_pot}
\end{figure}

\paragraph{Robosuite-Single task}
To demonstrate the efficiency of our proposed method, we first evaluate it under single-task setting in Robosuite, where we train an agent for each task.
Here, five pairs of tasks exhibiting time reversal symmetry are considered: door opening/closing inward, door opening/closing outward, peg insertion/removal, nut assembly/disassembly and block stacking/unstacking.
The IQM of agent performance across 10 environments are shown in \Cref{fig:IQM_rev_aug_filter_rev_pot}, and full evaluation curves are provided in \Cref{fig:rev_aug_filter_rev_pot} due to page limit. 
The results demonstrate the performance gain of our method over the baseline and confirm the contribution of each component in our method. 

\paragraph{Robosuite-Multi task}
We further evaluate our method in multi-task settings.
Our method is orthogonal to existing multi-task learning frameworks, meaning it can be seamlessly integrated with them. 
To highlight the performance gains of our approach, we demonstrate its effectiveness by combining it with existing multi-task methods.
Here, we start with training one agent for a pair of tasks.
Later on, we extend our method to using a single agent for all concerned tasks.
To train a single agent for multiple tasks, we consider extend the models to either taking an additional task embedding as input (task-conditioned) or outputting the actions of several tasks at the same time (multi-head).

\begin{figure}[th]
    \begin{minipage}{0.28\textwidth}
        \centering
        \includegraphics[width=0.99\linewidth]{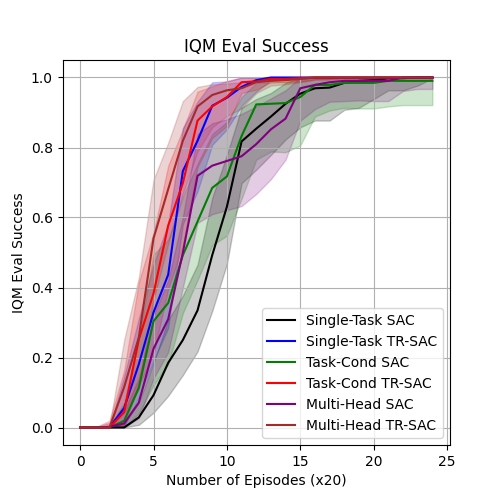}
    \end{minipage}
    \hfill
    \begin{minipage}{0.7\textwidth}
        \centering
        \small
        \begin{tabular}{c c c c c}
            \toprule
               \multirow{2}{*}{Method} & \multicolumn{4}{c}{Number of Environment Transitions} \\
                & 50k & 100k & 150k & 200k\\
            \midrule
                Single-Task SAC & \makecell{0.11$\pm$0.07} & \makecell{0.62$\pm$0.15} & \makecell{0.93$\pm$0.07} & \makecell{0.97$\pm$0.03} \\
            \midrule
                Task-Cond SAC & \makecell{0.31$\pm$0.19} & \makecell{0.71$\pm$0.16} & \makecell{0.90$\pm$0.10} & \makecell{0.96$\pm$0.04} \\
            \midrule
                Multi-Head SAC & \makecell{0.23$\pm$0.09} & \makecell{0.75$\pm$0.13} & \makecell{0.94$\pm$0.06} & \makecell{0.97$\pm$0.03} \\
            \midrule
                \makecell{Single-Task\\ TR-SAC (Ours)} & \makecell{0.33$\pm$0.14} & \makecell{0.92$\pm$0.07} & \textbf{\makecell{1.00$\pm$0.00}} & \textbf{\makecell{1.00$\pm$0.00}} \\
            \midrule
                \makecell{Task-Cond\\ TR-SAC (Ours)} & \makecell{0.38$\pm$0.17} & \makecell{0.93$\pm$0.07} & \makecell{0.99$\pm$0.01} & \textbf{\makecell{1.00$\pm$0.00}} \\
            \midrule
                \makecell{Multi-Head\\ TR-SAC (Ours)} & \textbf{\makecell{0.54$\pm$0.17}} & \textbf{\makecell{0.94$\pm$0.05}} & \makecell{0.99$\pm$0.01} & \textbf{\makecell{1.00$\pm$0.00}} \\
            \bottomrule
        \end{tabular}
    \end{minipage}
    \caption{
    \textbf{IQM of success rate for multi-task settings in 10 environments from Robosuite.}
    "Task-Cond" and "Multi-Head" are short for "task-conditioned" and "multi-headed" respectively.}
    \label{fig:IQM_robosuite10}
\end{figure}

For task-conditioned setting with only two tasks, we use one-hot encoding for the task embedding.
The actor, critic and potential models take both the state and the task embedding as input.
Considering that the environment dynamics are identical within each task pair, the pair of tasks share the forward and inverse dynamics models.
For multi-headed setting with only two tasks, the models output values for both two tasks simultaneously.
The performance of integrating our proposed method into these baselines in 10 environments of Robosuite is shown in \Cref{fig:IQM_robosuite10}.
The full evaluation curves of agent performance are included in \Cref{fig:robosuite_mt}.
Our proposed techniques clearly enhance the sample efficiency and improve the final performance when combined with the three baselines.

\begin{figure}[th]
    \begin{minipage}{0.35\textwidth}
        \centering
        \includegraphics[width=0.99\linewidth]{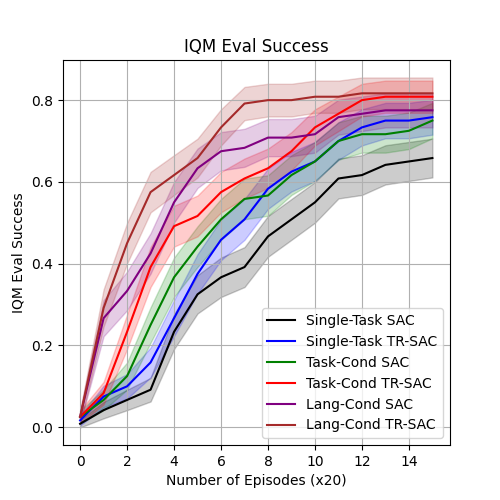}
    \end{minipage}
    \hfill
    \begin{minipage}{0.64\textwidth}
        \centering
        \small
        \begin{tabular}{c c c c}
        \toprule
           \multirow{2}{*}{Method} & \multicolumn{3}{c}{Number of Environment Transitions} \\
            & 50k & 100k & 150k\\
        \midrule
            Single-Task SAC & \makecell{0.33$\pm$0.05} & \makecell{0.55$\pm$0.05} & \makecell{0.66$\pm$0.05} \\
        \midrule
            Task-Cond SAC & \makecell{0.44$\pm$0.05} & \makecell{0.65$\pm$0.05} & \makecell{0.75$\pm$0.04}\\
        \midrule
            Lang-Cond SAC & \makecell{0.63$\pm$0.05} & \makecell{0.72$\pm$0.05} & \makecell{0.78$\pm$0.04}\\
        \midrule
            \makecell{Single-Task\\ TR-SAC (Ours)} & \makecell{0.38$\pm$0.05} & \makecell{0.65$\pm$0.05} & \makecell{0.76$\pm$0.04}\\
        \midrule
            \makecell{Task-Cond\\ TR-SAC (Ours)} & \makecell{0.52$\pm$0.05} & \makecell{0.73$\pm$0.04} & \makecell{0.81$\pm$0.04}\\
        \midrule
            \makecell{Lang-Cond\\ TR-SAC (Ours)} & \textbf{\makecell{0.66$\pm$0.05}} & \textbf{\makecell{0.81$\pm$0.04}} & \textbf{\makecell{0.82$\pm$0.04}}\\
        \bottomrule
        \end{tabular}
        \label{tab:IQM_reversible24}
    \end{minipage}   
    \caption{\textbf{IQM of success rate for multi-task settings in 12 pair of reversible tasks in MT50 of Meta-World.}
    "Task-Cond" and "Lang-Cond" are short for "task-conditioned" and "language-conditioned" respectively.}
    \label{fig:IQM_reversible24}
\end{figure}

\paragraph{Metaworld-Multi task}
Furthermore, we evaluate our methods on MT50, a benchmark with 50 environments from Meta-World.
Within the 50 tasks, we identify 12 pairs of reversible taks  and apply our techniques to these pairs.
Here, considering the exploding output dimensions when using multi-head setting for 50 tasks, we remove this baseline.
Instead, we introduce another baseline called language-conditioned SAC.
Here, the task embeddings are obtained by applying a pretrained language encoder, called CLIP \citep{radford2021learningtransferablevisualmodels}, on the language instructions of these tasks.
The IQM results for these 12 task pairs are shown in the right of \Cref{fig:IQM_reversible24}, and additional results including the average number of training episodes required to achieve a $100\%$ success rate are included in \Cref{tab:num_success_episode_metworld_reversible24} and \Cref{fig:num_success_episode_metworld_reversible24}.
We also present results for all 50 environments of MT50 in \Cref{fig:IQM_metaworld_MT50} and \Cref{fig:num_success_episode_metworld_MT50}.
With our proposed techniques, the agent learns faster and performs better compared to the baselines in both reversible tasks and all tasks of MT50.

\subsection{Ablation study}

\paragraph{Trajectory reversal augmentation with dynamics-aware filtering}
We analyze trajectory reversal augmentation with dynamics-aware filtering on three pairs of tasks: door opening/closing inward, door opening/closing outward, and peg insertion/removal.
As shown in \Cref{appendix: dynamics aware filter}, incorporating reversed transitions improves the performance for fully reversible tasks (e.g., door opening/closing outward and peg insertion/removal).
However, most transitions in door opening/closing inward are not fully reversible.
Including the reversed transitions generated by the inverse dynamics model leads to infeasible transitions, resulting in degraded performance.
After incorporating dynamics-aware filtering, which removes invalid reversed transitions, the performance surpasses the baseline for partially reversible tasks, demonstrating the effectiveness of our filtering strategy.
As for the hyperparameter $\beta$ that controls the filtering error tolerance, we finalize its value as $0.01$ after tuning among $[0.01, 0.001, 0.0001]$, with the related results presented in \Cref{appendix: beta_tune}.

\paragraph{Time reversal symmetry guided reward shaping}
Here we investigate the design choice for the time reversal symmetry-guided reward shaping.
We first explore how the potential models should be trained.
As shown in \Cref{appendix: tune_rev_pot}, it is concluded that two potential models should be trained with successful trajectories from the task itself, and from its reversible counterpart respectively.
Under this setting, the average of the rewards from these two models are used as the final reward. 
Moreover, four different types of potential value functions along the successful trajectories are compared.
The linear function outperforms the other choices, as shown in \Cref{appendix: pot_tune}.

\section{Conclusion}
We propose TR-DRL, a framework leveraging time reversal symmetry to enhance sample efficiency of DRL algorithms. 
Key contributions include a novel notion of partial time reversal symmetry, trajectory reversal augmentation with dynamics-aware filtering, and symmetry-guided reward shaping. 
Experiments on Robosuite and Metaworld demonstrate improved agent performance and learning efficiency. 
Future work may explore using prediction errors to identify reversible task pairs automatically, which allows deep reinforcement learning in robotics to be more efficient.




\clearpage

\section*{Acknowledgments} 
This work has been supported by the program of National Natural Science Foundation of China (No. 62176154), by the program of National Natural Science Foundation of China (No. 6250020129), and by Shanghai Magnolia Funding Pujiang Program (No. 23PJ1404400).

\bibliographystyle{unsrtnat}
\bibliography{mybibfile}

\begin{thebibliography}{31}
\providecommand{\natexlab}[1]{#1}
\providecommand{\url}[1]{\texttt{#1}}
\expandafter\ifx\csname urlstyle\endcsname\relax
  \providecommand{\doi}[1]{doi: #1}\else
  \providecommand{\doi}{doi: \begingroup \urlstyle{rm}\Url}\fi

\bibitem[Lin et~al.(2020)Lin, Huang, Zimmer, Guan, Rojas, and Weng]{Lin_2020}
Yijiong Lin, Jiancong Huang, Matthieu Zimmer, Yisheng Guan, Juan Rojas, and
  Paul Weng.
\newblock Invariant transform experience replay: Data augmentation for deep
  reinforcement learning.
\newblock \emph{IEEE Robotics and Automation Letters}, 5\penalty0 (4):\penalty0
  6615–6622, October 2020.
\newblock ISSN 2377-3774.
\newblock \doi{10.1109/lra.2020.3013937}.
\newblock URL \url{http://dx.doi.org/10.1109/LRA.2020.3013937}.

\bibitem[Kidzi{\'{n}}ski et~al.(2018)Kidzi{\'{n}}ski, Mohanty, Ong, Huang,
  Zhou, Pechenko, Stelmaszczyk, Jarosik, Pavlov, Kolesnikov, Plis, Chen, Zhang,
  Chen, Shi, Zheng, Yuan, Lin, Michalewski, Milos, Osinski, Melnik, Schilling,
  Ritter, Carroll, Hicks, Levine, Salath{\'e}, and Delp]{10.1007}
{\L}ukasz Kidzi{\'{n}}ski, Sharada~Prasanna Mohanty, Carmichael~F. Ong, Zhewei
  Huang, Shuchang Zhou, Anton Pechenko, Adam Stelmaszczyk, Piotr Jarosik,
  Mikhail Pavlov, Sergey Kolesnikov, Sergey Plis, Zhibo Chen, Zhizheng Zhang,
  Jiale Chen, Jun Shi, Zhuobin Zheng, Chun Yuan, Zhihui Lin, Henryk
  Michalewski, Piotr Milos, Blazej Osinski, Andrew Melnik, Malte Schilling,
  Helge Ritter, Sean~F. Carroll, Jennifer Hicks, Sergey Levine, Marcel
  Salath{\'e}, and Scott Delp.
\newblock Learning to run challenge solutions: Adapting reinforcement learning
  methods for neuromusculoskeletal environments.
\newblock In Sergio Escalera and Markus Weimer, editors, \emph{The NIPS '17
  Competition: Building Intelligent Systems}, pages 121--153, Cham, 2018.
  Springer International Publishing.
\newblock ISBN 978-3-319-94042-7.

\bibitem[Yarats et~al.(2022)Yarats, Fergus, Lazaric, and Pinto]{drqv2}
Denis Yarats, Rob Fergus, Alessandro Lazaric, and Lerrel Pinto.
\newblock Mastering visual continuous control: Improved data-augmented
  reinforcement learning.
\newblock In \emph{International Conference on Learning Representations}, 2022.
\newblock URL \url{https://openreview.net/forum?id=_SJ-_yyes8}.

\bibitem[Cohen and
  Welling(2016)]{cohen2016groupequivariantconvolutionalnetworks}
Taco~S. Cohen and Max Welling.
\newblock Group equivariant convolutional networks.
\newblock In \emph{Proceedings of the 33rd International Conference on
  International Conference on Machine Learning - Volume 48}, ICML'16, page
  2990–2999. JMLR.org, 2016.

\bibitem[Wang et~al.(2022)Wang, Walters, and
  Platt]{wang2022mathrmsoequivariant}
Dian Wang, Robin Walters, and Robert Platt.
\newblock \${\textbackslash}mathrm\{{SO}\}(2)\$-equivariant reinforcement
  learning.
\newblock In \emph{International Conference on Learning Representations}, 2022.
\newblock URL \url{https://openreview.net/forum?id=7F9cOhdvfk_}.

\bibitem[Hu et~al.(2024{\natexlab{a}})Hu, Jiang, and
  Weng]{hu2024revisitingdataaugmentationdeep}
Jianshu Hu, Yunpeng Jiang, and Paul Weng.
\newblock Revisiting data augmentation in deep reinforcement learning.
\newblock In \emph{The Twelfth International Conference on Learning
  Representations}, 2024{\natexlab{a}}.
\newblock URL \url{https://openreview.net/forum?id=EGQBpkIEuu}.

\bibitem[Raileanu et~al.(2021)Raileanu, Goldstein, Yarats, Kostrikov, and
  Fergus]{DrAC}
Roberta Raileanu, Maxwell Goldstein, Denis Yarats, Ilya Kostrikov, and Rob
  Fergus.
\newblock Automatic data augmentation for generalization in reinforcement
  learning.
\newblock In M.~Ranzato, A.~Beygelzimer, Y.~Dauphin, P.S. Liang, and J.~Wortman
  Vaughan, editors, \emph{Advances in Neural Information Processing Systems},
  volume~34, pages 5402--5415. Curran Associates, Inc., 2021.
\newblock URL
  \url{https://proceedings.neurips.cc/paper_files/paper/2021/file/2b38c2df6a49b97f706ec9148ce48d86-Paper.pdf}.

\bibitem[Hu et~al.(2024{\natexlab{b}})Hu, Weng, and
  Ban]{hu2024statenoveltyguidedactionpersistence}
Jianshu Hu, Paul Weng, and Yutong Ban.
\newblock State-novelty guided action persistence in deep reinforcement
  learning, 2024{\natexlab{b}}.
\newblock URL \url{https://arxiv.org/abs/2409.05433}.

\bibitem[Pitis et~al.(2020)Pitis, Creager, and
  Garg]{pitis2020counterfactualdataaugmentationusing}
Silviu Pitis, Elliot Creager, and Animesh Garg.
\newblock Counterfactual data augmentation using locally factored dynamics,
  2020.
\newblock URL \url{https://arxiv.org/abs/2007.02863}.

\bibitem[Corrado and
  Hanna(2024)]{corrado2024understandingdynamicsinvariantdataaugmentations}
Nicholas Corrado and Josiah~P. Hanna.
\newblock Understanding when dynamics-invariant data augmentations benefit
  model-free reinforcement learning updates.
\newblock In \emph{The Twelfth International Conference on Learning
  Representations}, 2024.
\newblock URL \url{https://openreview.net/forum?id=sVEu295o70}.

\bibitem[Corrado et~al.(2024)Corrado, Qu, Balis, Labiosa, and
  Hanna]{corrado2024guideddataaugmentationoffline}
Nicholas~E. Corrado, Yuxiao Qu, John~U. Balis, Adam Labiosa, and Josiah~P.
  Hanna.
\newblock Guided data augmentation for offline reinforcement learning and
  imitation learning.
\newblock In \emph{Reinforcement Learning Conference}, 2024.
\newblock URL \url{https://openreview.net/forum?id=rtJmC83c0r}.

\bibitem[Ma et~al.(2024)Ma, Wang, Yuan, Wang, Yuan, and Tao]{DA_DRL}
Guozheng Ma, Zhen Wang, Zhecheng Yuan, Xueqian Wang, Bo~Yuan, and Dacheng Tao.
\newblock A comprehensive survey of data augmentation in visual reinforcement
  learning, 2024.
\newblock URL \url{https://arxiv.org/abs/2210.04561}.

\bibitem[Sinha et~al.(2021)Sinha, Mandlekar, and
  Garg]{sinha2021s4rlsurprisinglysimpleselfsupervision}
Samarth Sinha, Ajay Mandlekar, and Animesh Garg.
\newblock S4rl: Surprisingly simple self-supervision for offline reinforcement
  learning, 2021.
\newblock URL \url{https://arxiv.org/abs/2103.06326}.

\bibitem[Qiao et~al.(2021)Qiao, Liang, Koltun, and
  Lin]{qiao2021efficientdifferentiablesimulationarticulated}
Yi-Ling Qiao, Junbang Liang, Vladlen Koltun, and Ming~C. Lin.
\newblock Efficient differentiable simulation of articulated bodies, 2021.
\newblock URL \url{https://arxiv.org/abs/2109.07719}.

\bibitem[Wang et~al.(2023)Wang, Park, Sortur, Wong, Walters, and
  Platt]{wang2023surprisingeffectivenessequivariantmodels}
Dian Wang, Jung~Yeon Park, Neel Sortur, Lawson~L.S. Wong, Robin Walters, and
  Robert Platt.
\newblock The surprising effectiveness of equivariant models in domains with
  latent symmetry.
\newblock In \emph{The Eleventh International Conference on Learning
  Representations}, 2023.
\newblock URL \url{https://openreview.net/forum?id=P4MUGRM4Acu}.

\bibitem[Barkley et~al.(2023)Barkley, Zhang, and
  Fridovich-Keil]{barkley2023investigationtimereversalsymmetry}
Brett Barkley, Amy Zhang, and David Fridovich-Keil.
\newblock An investigation of time reversal symmetry in reinforcement learning.
\newblock In \emph{Conference on Learning for Dynamics \& Control}, 2023.
\newblock URL \url{https://api.semanticscholar.org/CorpusID:265466253}.

\bibitem[Cheng et~al.(2023)Cheng, Zhan, Wu, Zhang, Lin, cheng Song, Wang, and
  Jiang]{cheng2023lookbeneathsurfaceexploiting}
Peng Cheng, Xianyuan Zhan, Zhihao Wu, Wenjia Zhang, Youfang Lin, Shou cheng
  Song, Han Wang, and Li~Jiang.
\newblock Look beneath the surface: Exploiting fundamental symmetry for
  sample-efficient offline {RL}.
\newblock In \emph{Thirty-seventh Conference on Neural Information Processing
  Systems}, 2023.
\newblock URL \url{https://openreview.net/forum?id=kyXMU3H7RB}.

\bibitem[Yao et~al.(2023)Yao, Bing, Zhuang, Chen, Zhou, Huang, and
  Knoll]{yao2023learningsymmetrymetareinforcementlearning}
Xiangtong Yao, Zhenshan Bing, Genghang Zhuang, Kejia Chen, Hongkuan Zhou, Kai
  Huang, and Alois Knoll.
\newblock Learning from symmetry: Meta-reinforcement learning with symmetrical
  behaviors and language instructions, 2023.
\newblock URL \url{https://arxiv.org/abs/2209.10656}.

\bibitem[Edwards et~al.(2018)Edwards, Downs, and
  Davidson]{edwards2018forwardbackwardreinforcementlearning}
Ashley~D. Edwards, Laura Downs, and James~C. Davidson.
\newblock Forward-backward reinforcement learning, 2018.
\newblock URL \url{https://arxiv.org/abs/1803.10227}.

\bibitem[Goyal et~al.(2019)Goyal, Brakel, Fedus, Singhal, Lillicrap, Levine,
  Larochelle, and Bengio]{goyal2019recalltracesbacktrackingmodels}
Anirudh Goyal, Philemon Brakel, William Fedus, Soumye Singhal, Timothy
  Lillicrap, Sergey Levine, Hugo Larochelle, and Yoshua Bengio.
\newblock Recall traces: Backtracking models for efficient reinforcement
  learning.
\newblock In \emph{International Conference on Learning Representations}, 2019.
\newblock URL \url{https://openreview.net/forum?id=HygsfnR9Ym}.

\bibitem[Nair et~al.(2020)Nair, Babaeizadeh, Finn, Levine, and
  Kumar]{nair2020timereversalselfsupervision}
Suraj Nair, Mohammad Babaeizadeh, Chelsea Finn, Sergey Levine, and Vikash
  Kumar.
\newblock Trass: Time reversal as self-supervision.
\newblock In \emph{2020 IEEE International Conference on Robotics and
  Automation (ICRA)}, pages 115--121, 2020.
\newblock \doi{10.1109/ICRA40945.2020.9196862}.

\bibitem[Grinsztajn et~al.(2021)Grinsztajn, Ferret, Pietquin, Preux, and
  Geist]{grinsztajn2021turningbackselfsupervisedapproach}
Nathan Grinsztajn, Johan Ferret, Olivier Pietquin, Philippe Preux, and Matthieu
  Geist.
\newblock There is no turning back: A self-supervised approach for
  reversibility-aware reinforcement learning.
\newblock In A.~Beygelzimer, Y.~Dauphin, P.~Liang, and J.~Wortman Vaughan,
  editors, \emph{Advances in Neural Information Processing Systems}, 2021.
\newblock URL \url{https://openreview.net/forum?id=3X65eaS4PtP}.

\bibitem[Eysenbach et~al.(2018)Eysenbach, Gu, Ibarz, and
  Levine]{eysenbach2017leavetracelearningreset}
Benjamin Eysenbach, Shixiang Gu, Julian Ibarz, and Sergey Levine.
\newblock Leave no trace: Learning to reset for safe and autonomous
  reinforcement learning.
\newblock In \emph{International Conference on Learning Representations}, 2018.
\newblock URL \url{https://openreview.net/forum?id=S1vuO-bCW}.

\bibitem[Ibrahim et~al.(2024)Ibrahim, Mostafa, Jnadi, Salloum, and
  Osinenko]{ibrahim2024comprehensiveoverviewrewardengineering}
Sinan Ibrahim, Mostafa Mostafa, Ali Jnadi, Hadi Salloum, and Pavel Osinenko.
\newblock Comprehensive overview of reward engineering and shaping in advancing
  reinforcement learning applications.
\newblock \emph{IEEE Access}, 12:\penalty0 175473--175500, 2024.
\newblock \doi{10.1109/ACCESS.2024.3504735}.

\bibitem[Ng et~al.(1999)Ng, Harada, and Russell]{10.5555/645528.657613}
Andrew~Y. Ng, Daishi Harada, and Stuart~J. Russell.
\newblock Policy invariance under reward transformations: Theory and
  application to reward shaping.
\newblock In \emph{Proceedings of the Sixteenth International Conference on
  Machine Learning}, ICML '99, page 278–287, San Francisco, CA, USA, 1999.
  Morgan Kaufmann Publishers Inc.
\newblock ISBN 1558606122.

\bibitem[Haarnoja et~al.(2018)Haarnoja, Zhou, Abbeel, and
  Levine]{haarnoja2018softactorcriticoffpolicymaximum}
Tuomas Haarnoja, Aurick Zhou, Pieter Abbeel, and Sergey Levine.
\newblock Soft actor-critic: Off-policy maximum entropy deep reinforcement
  learning with a stochastic actor.
\newblock In Jennifer Dy and Andreas Krause, editors, \emph{Proceedings of the
  35th International Conference on Machine Learning}, volume~80 of
  \emph{Proceedings of Machine Learning Research}, pages 1861--1870. PMLR,
  10--15 Jul 2018.
\newblock URL \url{https://proceedings.mlr.press/v80/haarnoja18b.html}.

\bibitem[Yu et~al.(2020)Yu, Quillen, He, Julian, Hausman, Finn, and
  Levine]{yu2021metaworldbenchmarkevaluationmultitask}
Tianhe Yu, Deirdre Quillen, Zhanpeng He, Ryan Julian, Karol Hausman, Chelsea
  Finn, and Sergey Levine.
\newblock Meta-world: A benchmark and evaluation for multi-task and meta
  reinforcement learning.
\newblock In Leslie~Pack Kaelbling, Danica Kragic, and Komei Sugiura, editors,
  \emph{Proceedings of the Conference on Robot Learning}, volume 100 of
  \emph{Proceedings of Machine Learning Research}, pages 1094--1100. PMLR, 30
  Oct--01 Nov 2020.
\newblock URL \url{https://proceedings.mlr.press/v100/yu20a.html}.

\bibitem[Zhu et~al.(2025)Zhu, Wong, Mandlekar, Martín-Martín, Joshi, Lin,
  Maddukuri, Nasiriany, and Zhu]{zhu2025robosuitemodularsimulationframework}
Yuke Zhu, Josiah Wong, Ajay Mandlekar, Roberto Martín-Martín, Abhishek Joshi,
  Kevin Lin, Abhiram Maddukuri, Soroush Nasiriany, and Yifeng Zhu.
\newblock robosuite: A modular simulation framework and benchmark for robot
  learning, 2025.
\newblock URL \url{https://arxiv.org/abs/2009.12293}.

\bibitem[Agarwal et~al.(2021)Agarwal, Schwarzer, Castro, Courville, and
  Bellemare]{agarwal2021deep}
Rishabh Agarwal, Max Schwarzer, Pablo~Samuel Castro, Aaron Courville, and
  Marc~G Bellemare.
\newblock Deep reinforcement learning at the edge of the statistical precipice.
\newblock \emph{Advances in Neural Information Processing Systems}, 2021.

\bibitem[Radford et~al.(2021)Radford, Kim, Hallacy, Ramesh, Goh, Agarwal,
  Sastry, Askell, Mishkin, Clark, Krueger, and
  Sutskever]{radford2021learningtransferablevisualmodels}
Alec Radford, Jong~Wook Kim, Chris Hallacy, Aditya Ramesh, Gabriel Goh,
  Sandhini Agarwal, Girish Sastry, Amanda Askell, Pamela Mishkin, Jack Clark,
  Gretchen Krueger, and Ilya Sutskever.
\newblock Learning transferable visual models from natural language
  supervision.
\newblock In Marina Meila and Tong Zhang, editors, \emph{Proceedings of the
  38th International Conference on Machine Learning}, volume 139 of
  \emph{Proceedings of Machine Learning Research}, pages 8748--8763. PMLR,
  18--24 Jul 2021.
\newblock URL \url{https://proceedings.mlr.press/v139/radford21a.html}.

\bibitem[Hendawy et~al.(2024)Hendawy, Peters, and D'Eramo]{hendawy2024multi}
Ahmed Hendawy, Jan Peters, and Carlo D'Eramo.
\newblock Multi-task reinforcement learning with mixture of orthogonal experts.
\newblock In \emph{Twelfth International Conference on Learning Representations
  (ICLR)}, 2024.
\newblock URL \url{https://arxiv.org/abs/2311.11385}.

\end{thebibliography}

\newpage
\section*{NeurIPS Paper Checklist}

\begin{enumerate}

\item {\bf Claims}
    \item[] Question: Do the main claims made in the abstract and introduction accurately reflect the paper's contributions and scope?
    \item[] Answer: \answerYes{} 
    \item[] Justification: We introduce our first and second contributions in \Cref{sec:method} and the third contribution in \Cref{sec: experimental results}.
    \item[] Guidelines:
    \begin{itemize}
        \item The answer NA means that the abstract and introduction do not include the claims made in the paper.
        \item The abstract and/or introduction should clearly state the claims made, including the contributions made in the paper and important assumptions and limitations. A No or NA answer to this question will not be perceived well by the reviewers. 
        \item The claims made should match theoretical and experimental results, and reflect how much the results can be expected to generalize to other settings. 
        \item It is fine to include aspirational goals as motivation as long as it is clear that these goals are not attained by the paper. 
    \end{itemize}

\item {\bf Limitations}
    \item[] Question: Does the paper discuss the limitations of the work performed by the authors?
    \item[] Answer: \answerYes{} 
    \item[] Justification: We discuss the limitations of our work in \Cref{appendix: limitations}.
    \item[] Guidelines:
    \begin{itemize}
        \item The answer NA means that the paper has no limitation while the answer No means that the paper has limitations, but those are not discussed in the paper. 
        \item The authors are encouraged to create a separate "Limitations" section in their paper.
        \item The paper should point out any strong assumptions and how robust the results are to violations of these assumptions (e.g., independence assumptions, noiseless settings, model well-specification, asymptotic approximations only holding locally). The authors should reflect on how these assumptions might be violated in practice and what the implications would be.
        \item The authors should reflect on the scope of the claims made, e.g., if the approach was only tested on a few datasets or with a few runs. In general, empirical results often depend on implicit assumptions, which should be articulated.
        \item The authors should reflect on the factors that influence the performance of the approach. For example, a facial recognition algorithm may perform poorly when image resolution is low or images are taken in low lighting. Or a speech-to-text system might not be used reliably to provide closed captions for online lectures because it fails to handle technical jargon.
        \item The authors should discuss the computational efficiency of the proposed algorithms and how they scale with dataset size.
        \item If applicable, the authors should discuss possible limitations of their approach to address problems of privacy and fairness.
        \item While the authors might fear that complete honesty about limitations might be used by reviewers as grounds for rejection, a worse outcome might be that reviewers discover limitations that aren't acknowledged in the paper. The authors should use their best judgment and recognize that individual actions in favor of transparency play an important role in developing norms that preserve the integrity of the community. Reviewers will be specifically instructed to not penalize honesty concerning limitations.
    \end{itemize}

\item {\bf Theory assumptions and proofs}
    \item[] Question: For each theoretical result, does the paper provide the full set of assumptions and a complete (and correct) proof?
    \item[] Answer: \answerNA{} 
    \item[] Justification: The paper does not include theoretical results. 
    \item[] Guidelines:
    \begin{itemize}
        \item The answer NA means that the paper does not include theoretical results. 
        \item All the theorems, formulas, and proofs in the paper should be numbered and cross-referenced.
        \item All assumptions should be clearly stated or referenced in the statement of any theorems.
        \item The proofs can either appear in the main paper or the supplemental material, but if they appear in the supplemental material, the authors are encouraged to provide a short proof sketch to provide intuition. 
        \item Inversely, any informal proof provided in the core of the paper should be complemented by formal proofs provided in appendix or supplemental material.
        \item Theorems and Lemmas that the proof relies upon should be properly referenced. 
    \end{itemize}

    \item {\bf Experimental result reproducibility}
    \item[] Question: Does the paper fully disclose all the information needed to reproduce the main experimental results of the paper to the extent that it affects the main claims and/or conclusions of the paper (regardless of whether the code and data are provided or not)?
    \item[] Answer: \answerYes{} 
    \item[] Justification: We include the pseudocode of our proposed method and all hyperparameters required to reproduce our experimental results in \Cref{alg:pseudocode} and \Cref{appendix: hyperparameter}.
    \item[] Guidelines:
    \begin{itemize}
        \item The answer NA means that the paper does not include experiments.
        \item If the paper includes experiments, a No answer to this question will not be perceived well by the reviewers: Making the paper reproducible is important, regardless of whether the code and data are provided or not.
        \item If the contribution is a dataset and/or model, the authors should describe the steps taken to make their results reproducible or verifiable. 
        \item Depending on the contribution, reproducibility can be accomplished in various ways. For example, if the contribution is a novel architecture, describing the architecture fully might suffice, or if the contribution is a specific model and empirical evaluation, it may be necessary to either make it possible for others to replicate the model with the same dataset, or provide access to the model. In general. releasing code and data is often one good way to accomplish this, but reproducibility can also be provided via detailed instructions for how to replicate the results, access to a hosted model (e.g., in the case of a large language model), releasing of a model checkpoint, or other means that are appropriate to the research performed.
        \item While NeurIPS does not require releasing code, the conference does require all submissions to provide some reasonable avenue for reproducibility, which may depend on the nature of the contribution. For example
        \begin{enumerate}
            \item If the contribution is primarily a new algorithm, the paper should make it clear how to reproduce that algorithm.
            \item If the contribution is primarily a new model architecture, the paper should describe the architecture clearly and fully.
            \item If the contribution is a new model (e.g., a large language model), then there should either be a way to access this model for reproducing the results or a way to reproduce the model (e.g., with an open-source dataset or instructions for how to construct the dataset).
            \item We recognize that reproducibility may be tricky in some cases, in which case authors are welcome to describe the particular way they provide for reproducibility. In the case of closed-source models, it may be that access to the model is limited in some way (e.g., to registered users), but it should be possible for other researchers to have some path to reproducing or verifying the results.
        \end{enumerate}
    \end{itemize}

\item {\bf Open access to data and code}
    \item[] Question: Does the paper provide open access to the data and code, with sufficient instructions to faithfully reproduce the main experimental results, as described in supplemental material?
    \item[] Answer: \answerYes{} 
    \item[] Justification: In the camera-ready version, we include the link to our source code repository.
    \item[] Guidelines:
    \begin{itemize}
        \item The answer NA means that paper does not include experiments requiring code.
        \item Please see the NeurIPS code and data submission guidelines (\url{https://nips.cc/public/guides/CodeSubmissionPolicy}) for more details.
        \item While we encourage the release of code and data, we understand that this might not be possible, so “No” is an acceptable answer. Papers cannot be rejected simply for not including code, unless this is central to the contribution (e.g., for a new open-source benchmark).
        \item The instructions should contain the exact command and environment needed to run to reproduce the results. See the NeurIPS code and data submission guidelines (\url{https://nips.cc/public/guides/CodeSubmissionPolicy}) for more details.
        \item The authors should provide instructions on data access and preparation, including how to access the raw data, preprocessed data, intermediate data, and generated data, etc.
        \item The authors should provide scripts to reproduce all experimental results for the new proposed method and baselines. If only a subset of experiments are reproducible, they should state which ones are omitted from the script and why.
        \item At submission time, to preserve anonymity, the authors should release anonymized versions (if applicable).
        \item Providing as much information as possible in supplemental material (appended to the paper) is recommended, but including URLs to data and code is permitted.
    \end{itemize}

\item {\bf Experimental setting/details}
    \item[] Question: Does the paper specify all the training and test details (e.g., data splits, hyperparameters, how they were chosen, type of optimizer, etc.) necessary to understand the results?
    \item[] Answer: \answerYes{} 
    \item[] Justification: We introduce the environments that we use in \Cref{appendix: environments} and provide all hyperparameters in \Cref{appendix: hyperparameter}.
    \item[] Guidelines:
    \begin{itemize}
        \item The answer NA means that the paper does not include experiments.
        \item The experimental setting should be presented in the core of the paper to a level of detail that is necessary to appreciate the results and make sense of them.
        \item The full details can be provided either with the code, in appendix, or as supplemental material.
    \end{itemize}

\item {\bf Experiment statistical significance}
    \item[] Question: Does the paper report error bars suitably and correctly defined or other appropriate information about the statistical significance of the experiments?
    \item[] Answer: \answerYes{} 
    \item[] Justification: We provide inter-quartile mean (IQM) for aggregated performance with 95\% confidence interval.
    \item[] Guidelines:
    \begin{itemize}
        \item The answer NA means that the paper does not include experiments.
        \item The authors should answer "Yes" if the results are accompanied by error bars, confidence intervals, or statistical significance tests, at least for the experiments that support the main claims of the paper.
        \item The factors of variability that the error bars are capturing should be clearly stated (for example, train/test split, initialization, random drawing of some parameter, or overall run with given experimental conditions).
        \item The method for calculating the error bars should be explained (closed form formula, call to a library function, bootstrap, etc.)
        \item The assumptions made should be given (e.g., Normally distributed errors).
        \item It should be clear whether the error bar is the standard deviation or the standard error of the mean.
        \item It is OK to report 1-sigma error bars, but one should state it. The authors should preferably report a 2-sigma error bar than state that they have a 96\% CI, if the hypothesis of Normality of errors is not verified.
        \item For asymmetric distributions, the authors should be careful not to show in tables or figures symmetric error bars that would yield results that are out of range (e.g. negative error rates).
        \item If error bars are reported in tables or plots, The authors should explain in the text how they were calculated and reference the corresponding figures or tables in the text.
    \end{itemize}

\item {\bf Experiments compute resources}
    \item[] Question: For each experiment, does the paper provide sufficient information on the computer resources (type of compute workers, memory, time of execution) needed to reproduce the experiments?
    \item[] Answer: \answerYes{} 
    \item[] Justification: We have stated our compute resources in \Cref{appendix:compute}.
    \item[] Guidelines:
    \begin{itemize}
        \item The answer NA means that the paper does not include experiments.
        \item The paper should indicate the type of compute workers CPU or GPU, internal cluster, or cloud provider, including relevant memory and storage.
        \item The paper should provide the amount of compute required for each of the individual experimental runs as well as estimate the total compute. 
        \item The paper should disclose whether the full research project required more compute than the experiments reported in the paper (e.g., preliminary or failed experiments that didn't make it into the paper). 
    \end{itemize}
    
\item {\bf Code of ethics}
    \item[] Question: Does the research conducted in the paper conform, in every respect, with the NeurIPS Code of Ethics \url{https://neurips.cc/public/EthicsGuidelines}?
    \item[] Answer: \answerYes{} 
    \item[] Justification: We have read the NeurIPS Code of Ethics carefully and conformed with it.
    \item[] Guidelines:
    \begin{itemize}
        \item The answer NA means that the authors have not reviewed the NeurIPS Code of Ethics.
        \item If the authors answer No, they should explain the special circumstances that require a deviation from the Code of Ethics.
        \item The authors should make sure to preserve anonymity (e.g., if there is a special consideration due to laws or regulations in their jurisdiction).
    \end{itemize}

\item {\bf Broader impacts}
    \item[] Question: Does the paper discuss both potential positive societal impacts and negative societal impacts of the work performed?
    \item[] Answer: \answerYes{} 
    \item[] Justification: We have stated the broader impacts of our paper in \Cref{appendix: broader impacts}.
    \item[] Guidelines:
    \begin{itemize}
        \item The answer NA means that there is no societal impact of the work performed.
        \item If the authors answer NA or No, they should explain why their work has no societal impact or why the paper does not address societal impact.
        \item Examples of negative societal impacts include potential malicious or unintended uses (e.g., disinformation, generating fake profiles, surveillance), fairness considerations (e.g., deployment of technologies that could make decisions that unfairly impact specific groups), privacy considerations, and security considerations.
        \item The conference expects that many papers will be foundational research and not tied to particular applications, let alone deployments. However, if there is a direct path to any negative applications, the authors should point it out. For example, it is legitimate to point out that an improvement in the quality of generative models could be used to generate deepfakes for disinformation. On the other hand, it is not needed to point out that a generic algorithm for optimizing neural networks could enable people to train models that generate Deepfakes faster.
        \item The authors should consider possible harms that could arise when the technology is being used as intended and functioning correctly, harms that could arise when the technology is being used as intended but gives incorrect results, and harms following from (intentional or unintentional) misuse of the technology.
        \item If there are negative societal impacts, the authors could also discuss possible mitigation strategies (e.g., gated release of models, providing defenses in addition to attacks, mechanisms for monitoring misuse, mechanisms to monitor how a system learns from feedback over time, improving the efficiency and accessibility of ML).
    \end{itemize}
    
\item {\bf Safeguards}
    \item[] Question: Does the paper describe safeguards that have been put in place for responsible release of data or models that have a high risk for misuse (e.g., pretrained language models, image generators, or scraped datasets)?
    \item[] Answer: \answerNA{} 
    \item[] Justification: Our work has no such risks.
    \item[] Guidelines:
    \begin{itemize}
        \item The answer NA means that the paper poses no such risks.
        \item Released models that have a high risk for misuse or dual-use should be released with necessary safeguards to allow for controlled use of the model, for example by requiring that users adhere to usage guidelines or restrictions to access the model or implementing safety filters. 
        \item Datasets that have been scraped from the Internet could pose safety risks. The authors should describe how they avoided releasing unsafe images.
        \item We recognize that providing effective safeguards is challenging, and many papers do not require this, but we encourage authors to take this into account and make a best faith effort.
    \end{itemize}

\item {\bf Licenses for existing assets}
    \item[] Question: Are the creators or original owners of assets (e.g., code, data, models), used in the paper, properly credited and are the license and terms of use explicitly mentioned and properly respected?
    \item[] Answer: \answerYes{} 
    \item[] Justification:  In our work, we do experiments on robosuite and Meta-World and have cited them properly in the paragraph of experimental setup in \Cref{sec:experimental setup}.
    \item[] Guidelines:
    \begin{itemize}
        \item The answer NA means that the paper does not use existing assets.
        \item The authors should cite the original paper that produced the code package or dataset.
        \item The authors should state which version of the asset is used and, if possible, include a URL.
        \item The name of the license (e.g., CC-BY 4.0) should be included for each asset.
        \item For scraped data from a particular source (e.g., website), the copyright and terms of service of that source should be provided.
        \item If assets are released, the license, copyright information, and terms of use in the package should be provided. For popular datasets, \url{paperswithcode.com/datasets} has curated licenses for some datasets. Their licensing guide can help determine the license of a dataset.
        \item For existing datasets that are re-packaged, both the original license and the license of the derived asset (if it has changed) should be provided.
        \item If this information is not available online, the authors are encouraged to reach out to the asset's creators.
    \end{itemize}

\item {\bf New assets}
    \item[] Question: Are new assets introduced in the paper well documented and is the documentation provided alongside the assets?
    \item[] Answer: \answerNA{} 
    \item[] Justification: Our work does not release new assets.
    \item[] Guidelines:
    \begin{itemize}
        \item The answer NA means that the paper does not release new assets.
        \item Researchers should communicate the details of the dataset/code/model as part of their submissions via structured templates. This includes details about training, license, limitations, etc. 
        \item The paper should discuss whether and how consent was obtained from people whose asset is used.
        \item At submission time, remember to anonymize your assets (if applicable). You can either create an anonymized URL or include an anonymized zip file.
    \end{itemize}

\item {\bf Crowdsourcing and research with human subjects}
    \item[] Question: For crowdsourcing experiments and research with human subjects, does the paper include the full text of instructions given to participants and screenshots, if applicable, as well as details about compensation (if any)? 
    \item[] Answer: \answerNA{} 
    \item[] Justification: Our work does not involve crowdsourcing nor research with human subjects.
    \item[] Guidelines:
    \begin{itemize}
        \item The answer NA means that the paper does not involve crowdsourcing nor research with human subjects.
        \item Including this information in the supplemental material is fine, but if the main contribution of the paper involves human subjects, then as much detail as possible should be included in the main paper. 
        \item According to the NeurIPS Code of Ethics, workers involved in data collection, curation, or other labor should be paid at least the minimum wage in the country of the data collector. 
    \end{itemize}

\item {\bf Institutional review board (IRB) approvals or equivalent for research with human subjects}
    \item[] Question: Does the paper describe potential risks incurred by study participants, whether such risks were disclosed to the subjects, and whether Institutional Review Board (IRB) approvals (or an equivalent approval/review based on the requirements of your country or institution) were obtained?
    \item[] Answer: \answerNA{} 
    \item[] Justification: Our work does not involve crowdsourcing nor research with human subjects.
    \item[] Guidelines:
    \begin{itemize}
        \item The answer NA means that the paper does not involve crowdsourcing nor research with human subjects.
        \item Depending on the country in which research is conducted, IRB approval (or equivalent) may be required for any human subjects research. If you obtained IRB approval, you should clearly state this in the paper. 
        \item We recognize that the procedures for this may vary significantly between institutions and locations, and we expect authors to adhere to the NeurIPS Code of Ethics and the guidelines for their institution. 
        \item For initial submissions, do not include any information that would break anonymity (if applicable), such as the institution conducting the review.
    \end{itemize}

\item {\bf Declaration of LLM usage}
    \item[] Question: Does the paper describe the usage of LLMs if it is an important, original, or non-standard component of the core methods in this research? Note that if the LLM is used only for writing, editing, or formatting purposes and does not impact the core methodology, scientific rigorousness, or originality of the research, declaration is not required.
    \item[] Answer: \answerNA{} 
    \item[] Justification: The LLM is used only for writing, editing, or formatting purposes and does not impact the core methodology, scientific rigorousness, or originality of the research. 
    The core method development in our work does not involve LLMs as any important, original, or non-standard components.
    \item[] Guidelines:
    \begin{itemize}
        \item The answer NA means that the core method development in this research does not involve LLMs as any important, original, or non-standard components.
        \item Please refer to our LLM policy (\url{https://neurips.cc/Conferences/2025/LLM}) for what should or should not be described.
    \end{itemize}

\end{enumerate}


\appendix

\section{Environments}
\label{appendix: environments}
\begin{figure}
    \begin{minipage}{0.23\textwidth}
        \centering
        \includegraphics[width=0.99\linewidth]{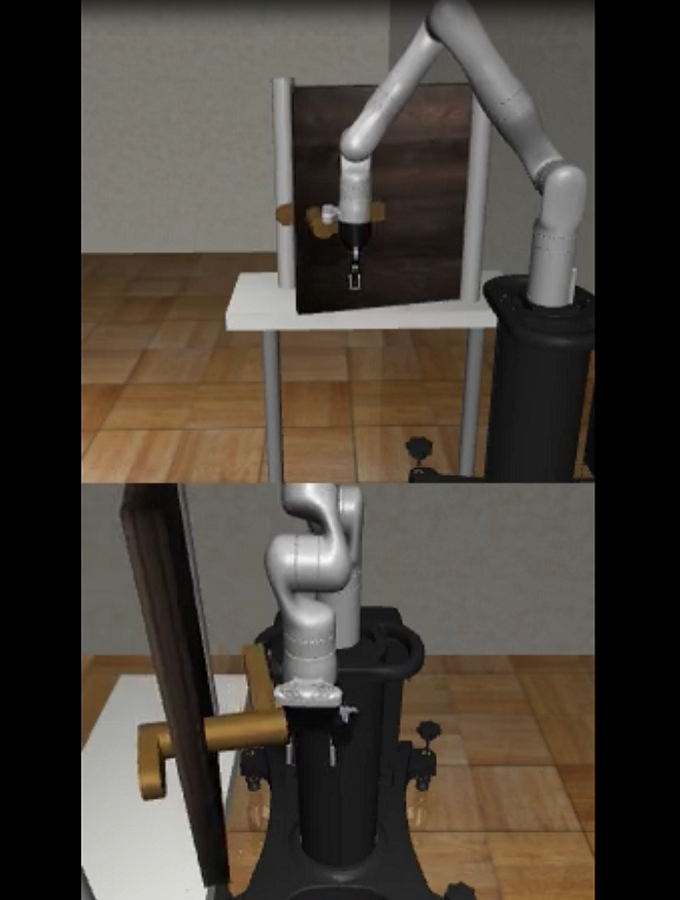} 
        \captionof{figure}{Door closing inward.}
        \label{fig:env_sample_door_close_inward}
    \end{minipage}
    \hfill
    \begin{minipage}{0.23\textwidth}
        \centering
        \includegraphics[width=0.99\linewidth]{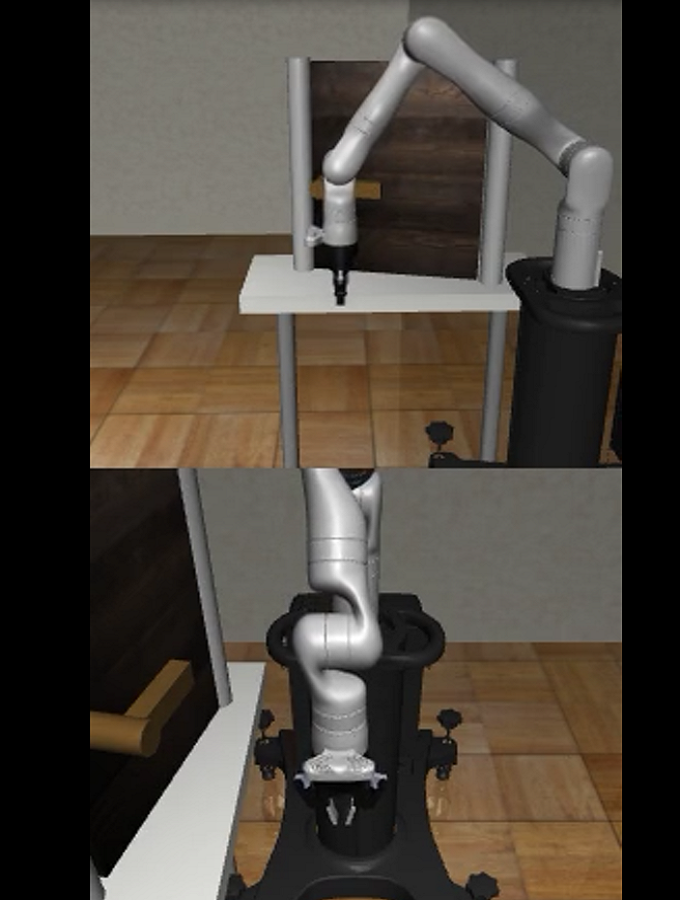}
        \captionof{figure}{Door opening inward.}
        \label{fig:env_sample_door_open_inward}
    \end{minipage}
    \hfill
    \begin{minipage}{0.23\textwidth}
        \centering
        \includegraphics[width=0.99\linewidth]{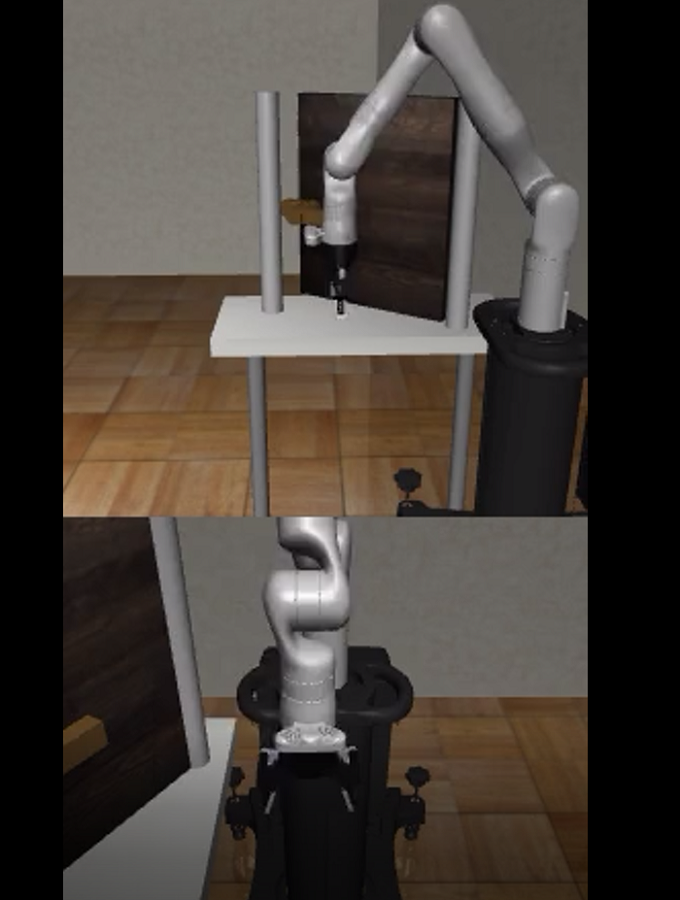} 
        \captionof{figure}{Door closing outward.}
        \label{fig:env_sample_door_close_outward}
    \end{minipage}
    \hfill
    \begin{minipage}{0.23\textwidth}
        \centering
        \includegraphics[width=0.99\linewidth]{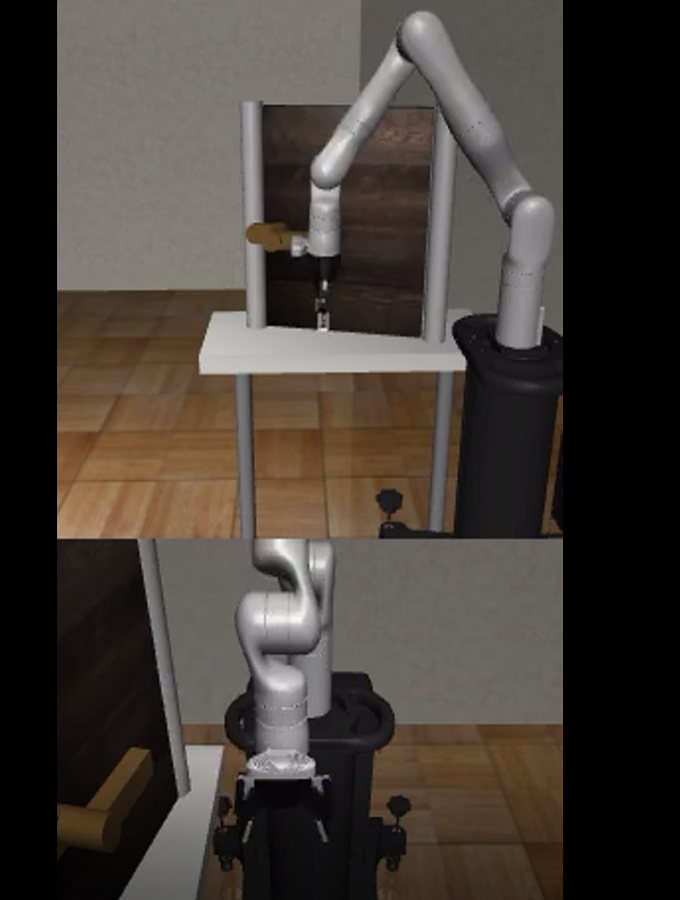}
        \captionof{figure}{Door opening outward.}
        \label{fig:env_sample_door_open_outward}
    \end{minipage}
\end{figure}

\begin{figure}
    \begin{minipage}{0.23\textwidth}
        \centering
        \includegraphics[width=0.99\linewidth]{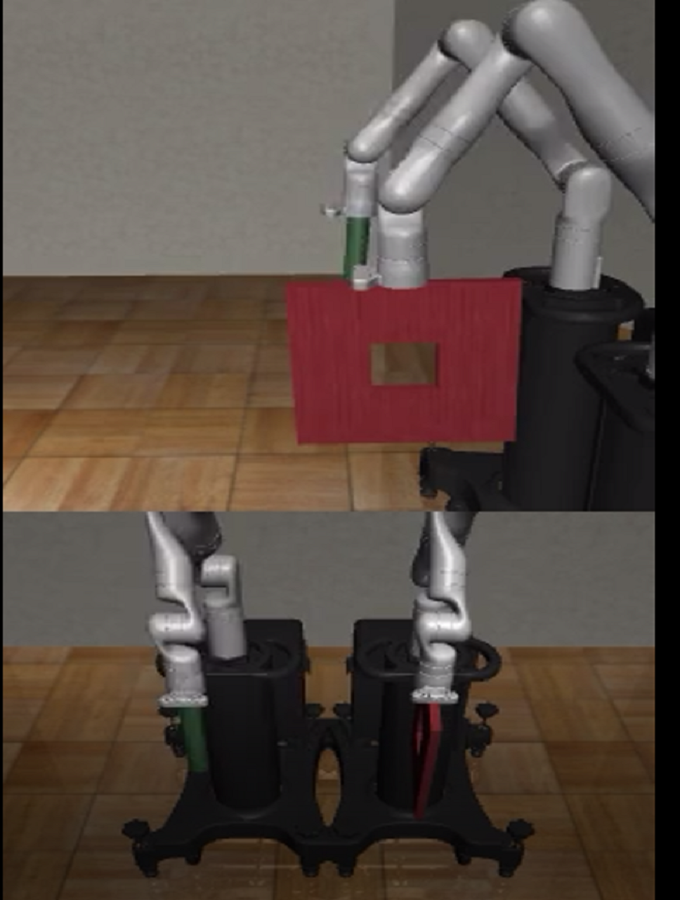} 
        \captionof{figure}{Peg insertion.}
        \label{fig:env_sample_peg_insertion}
    \end{minipage}
    \hfill
    \begin{minipage}{0.23\textwidth}
        \centering
        \includegraphics[width=0.99\linewidth]{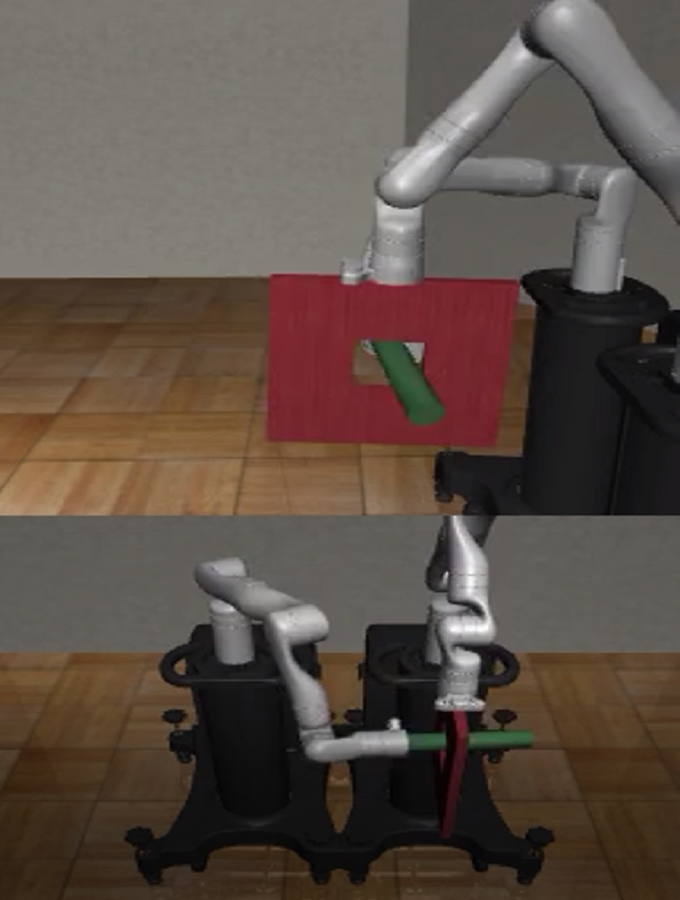}
        \captionof{figure}{Peg removal.}
        \label{fig:env_sample_peg_removal}
    \end{minipage}
    \hfill
    \begin{minipage}{0.23\textwidth}
        \centering
        \includegraphics[width=0.99\linewidth]{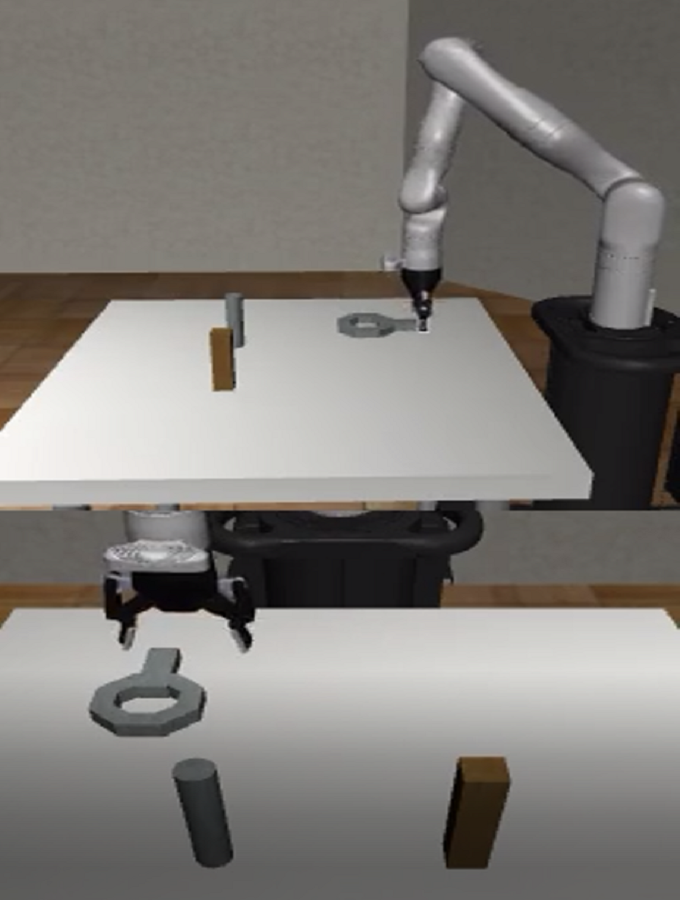} 
        \captionof{figure}{Nut assembly.}
        \label{fig:env_sample_nut_assembly}
    \end{minipage}
    \hfill
    \begin{minipage}{0.23\textwidth}
        \centering
        \includegraphics[width=0.99\linewidth]{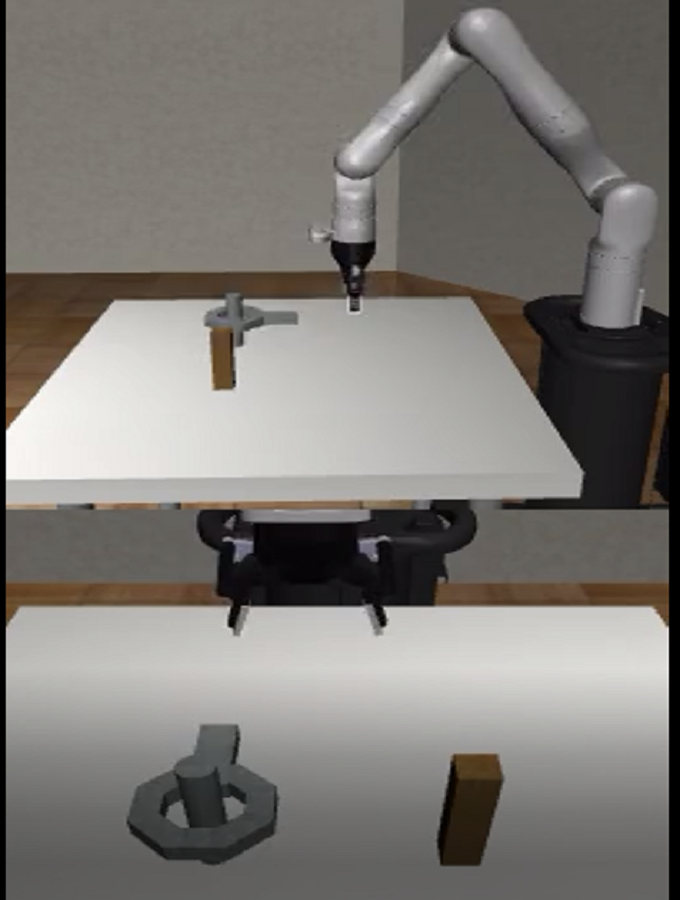}
        \captionof{figure}{Nut disassembly.}
        \label{fig:env_sample_nut_disassembly}
    \end{minipage}
\end{figure}

We introduce the environments utilized in our Robosuite experiments.
\begin{itemize}
    \item Door Opening/Closing Inward: 
    The agent needs to open/close the door inward.
    "Inward" means that the door is on the same side as the robotics arm.
    The agent can close the door by pushing it.
    To open the door, the agent has to grasp the handle and pull the handle to a desired position, making this task pair partially time reversal symmetric.
    Examples are shown in \Cref{fig:env_sample_door_close_inward} and \Cref{fig:env_sample_door_open_inward}.
    \item Door Opening/Closing Outward:
    The agent needs to open/close the door outward.
    "Outward" indicates that the door is on the opposite side as the robotics arm.
    In this task pair, the agent has to grasp the handle and then open/close the door, making this task pair fully time reversal symmetric.
    Examples are shown in \Cref{fig:env_sample_door_close_outward} and \Cref{fig:env_sample_door_open_outward}.
    \item Peg Insertion/Removal:
    The agent needs to insert/remove the peg into/out of the hole.
    Examples are shown in \Cref{fig:env_sample_peg_insertion} and \Cref{fig:env_sample_peg_removal}.
    \item Nut Assembly/Disassembly:
    The agent needs to assemble/disassemble the nut.
    Examples are shown in \Cref{fig:env_sample_nut_assembly} and \Cref{fig:env_sample_nut_disassembly}.
    \item Block Stack/Unstack.
    The agent needs to either stack a small block onto a larger one or unstack it by removing the small block.
\end{itemize}

The introductions of environments that we have used in our Meta-World experiments can be found in \citep{yu2021metaworldbenchmarkevaluationmultitask}, which also provides the language instruction for each task.

\section{Implementation Details}
\label{appendix: hyperparameter}
As shown in \Cref{tab:hyperparameters}, we present the value of hyperparameters used in our experiments. 
For experiments in Robosuite, we adopt the hyperparameters specified by \citet{haarnoja2018softactorcriticoffpolicymaximum}.
In the case of experiments on MT50 of Metaworld, we primarily follow the hyperparameter settings provided by \citet{yu2021metaworldbenchmarkevaluationmultitask}.
Furthermore, in our source code \ref{footnote: source code}, we include the implementation of our method built upon MOORE \citep{hendawy2024multi}, a more recent and advanced baseline compared to SAC for multi-task RL.

\begin{table}[ht]
    \centering
    \begin{tabular}{c c c}
    \toprule
       Hyperparameter & Robosuite & MetaWorld \\
    \midrule
       hidden depth & 2 & 3 \\
       hidden dimension & 512 & 400 \\
       horizon & 500 & 200 \\
       environment steps & 250,000 & 100,000 \\
       replay buffer capacity & 250,000 & 100,000 \\
       random steps & 5,000 & 2,000 \\
       batch size & 512 & 128 \\
       discount & 0.99 & 0.99\\
       learning rate & 1e-3 & 3e-4 \\
       learning rate ($\alpha$ of SAC) & 1e-3 & 3e-4 \\
       target network update frequency & 2 & 1 \\
       target network soft-update rate & 0.01 & 0.005 \\
       actor update frequency & 2 & 2 \\
       actor log stddev bounds & [-10, 2] & [-20, 2] \\
       init temperature & 0.1 & 0.1 \\
    \bottomrule
    \end{tabular}
    \caption{Hyperparameters used in our experiments.}
    \label{tab:hyperparameters}
\end{table}

\section{Ablation Study of Dynamics-Aware Filtering in Trajectory Reversal Augmentation}
\label{appendix: dynamics aware filter}
\begin{figure}
    \centering
    \includegraphics[width=0.9\linewidth]{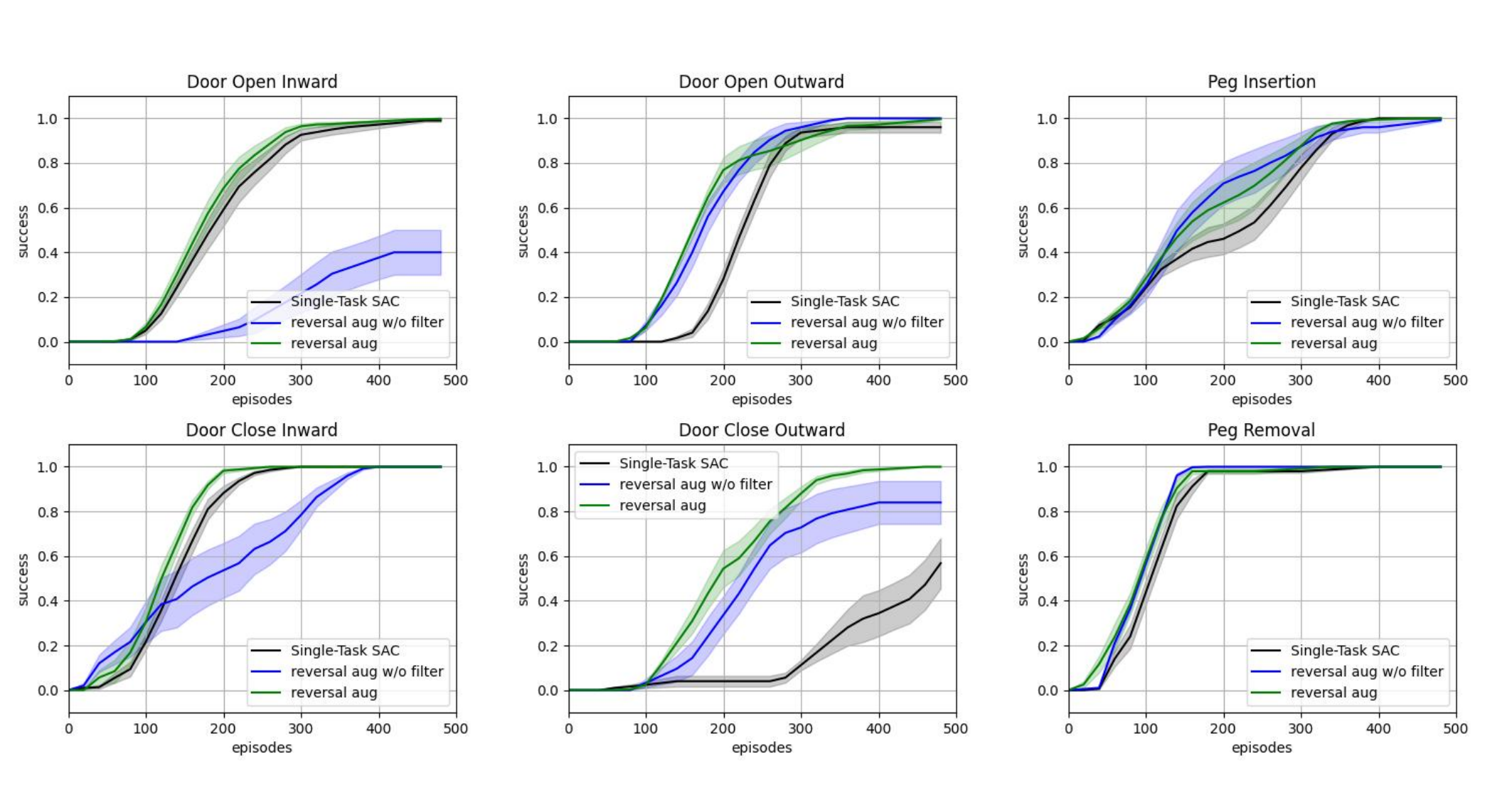}
    \caption{Evaluation curves of trajectory reversal augmentation with dynamics-aware filtering in 6 environments of robosuite. 
    "Single-Task SAC" serves as the baseline. 
    "+reversal aug w/o filter" introduces trajectory reversal augmentation without filtering, while "+reversal aug" incorporates dynamics-aware filtering for reversed transitions.}
    \label{fig:rev_aug_10demo}
\end{figure}

As shown in \Cref{fig:rev_aug_10demo}, incorporating reversed transitions improves the performance for fully reversible tasks (e.g., door opening/closing outward and peg insertion/removal).
However, most transitions in door opening/closing inward are not fully reversible.
Including the reversed transitions generated by the inverse dynamics model leads to infeasible transitions, resulting in degraded performance.
After incorporating dynamics-aware filtering, which removes invalid reversed transitions, the performance surpasses the baseline for partially reversible tasks, demonstrating the effectiveness of our filtering strategy.

\section{Hyperparameter Tuning in Dynamics-Aware Filtering}
\label{appendix: beta_tune}
\begin{figure}
    \centering
    \includegraphics[width=0.99\linewidth]{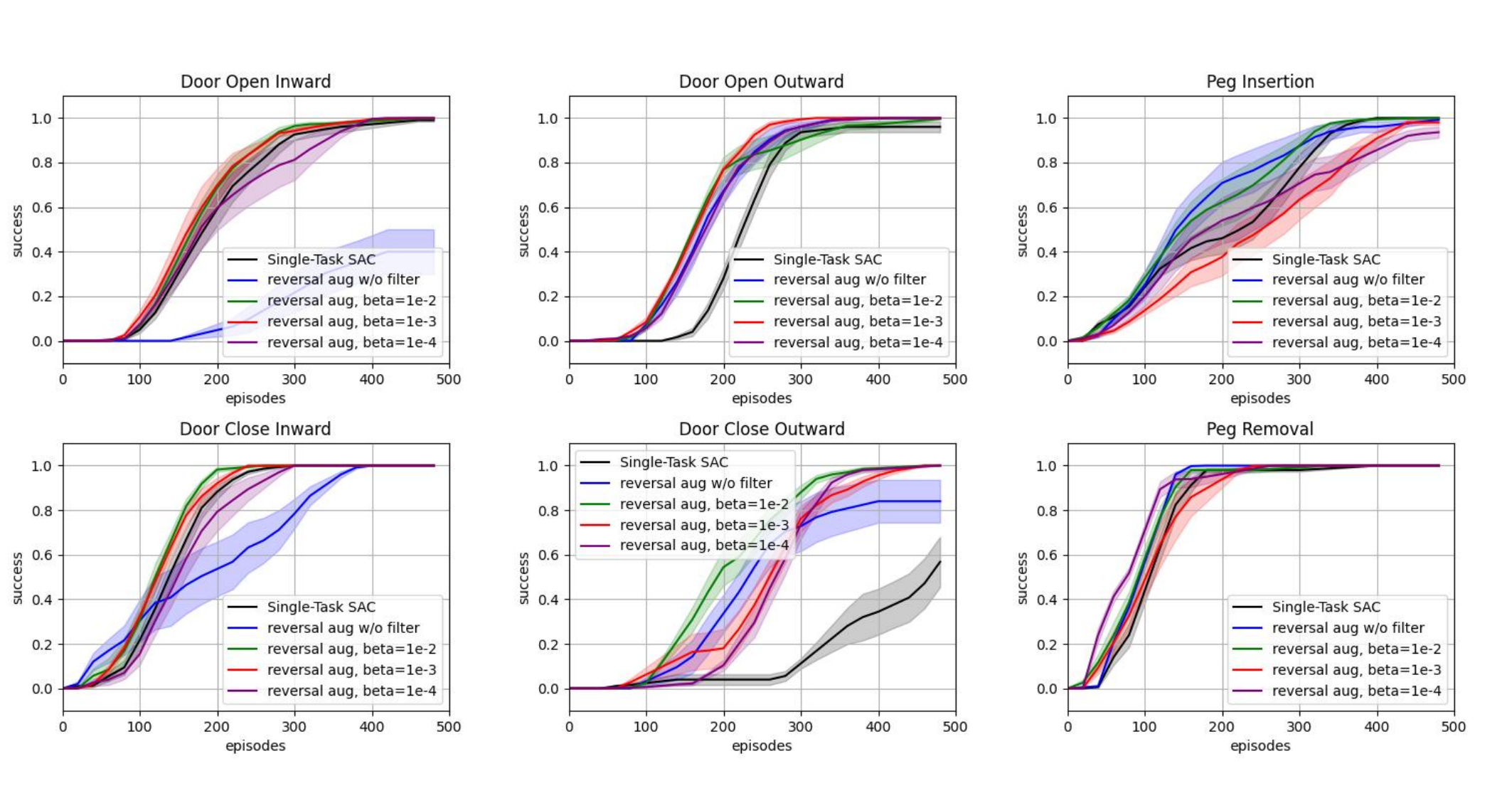}
    \caption{Evaluation curves of agent success rate using trajectory reversal augmentation with dynamics-aware filtering (different $\beta$).}
    \label{fig:rev_aug_10demo_compare_beta}
\end{figure}

\begin{figure}
    \centering
    \includegraphics[width=0.35\linewidth]{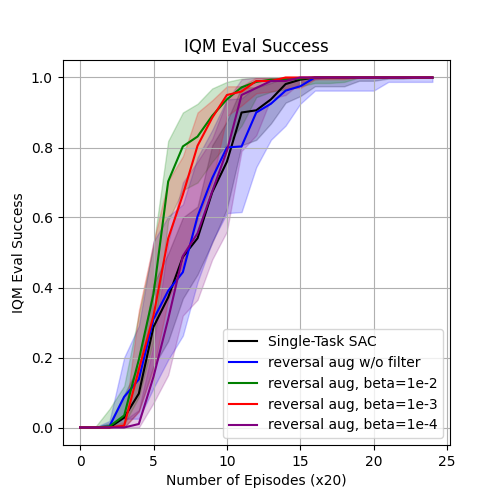}
    \caption{IQM of agent success rate using trajectory reversal augmentation with dynamics-aware filtering with different $\beta$.}
    \label{fig:IQM_rev_aug_10demo_compare_beta}
\end{figure}

Here, we perform hyperparameter tuning of $\beta$ in dynamics-aware filtering.
Recall that $\beta$ governs the error tolerance for reversed transitions: $\beta=0.01$ filters out transitions where $\mid\mid s - \hat{s} \mid\mid \geq 0.01\cdot||s_{\max} - s_{\min}||$, where $s_{\max}$ and $s_{\min}$ represent the state space extremes.
We conduct a linear search for $\beta$ among $[0.01, 0.001, 0.0001]$.
Evaluation curves of agent success rate in these six environments are shown in \Cref{fig:rev_aug_10demo_compare_beta}, while the inter-quartile mean of agent success rate is presented in \Cref{fig:IQM_rev_aug_10demo_compare_beta}.
Based on these results, we select $\beta=0.01$ for subsequent experiments.

\section{Separate Plots of \Cref{fig:IQM_robosuite10} and \Cref{fig:IQM_reversible24}}
\begin{figure}
    \centering
    \includegraphics[width=0.9\linewidth]{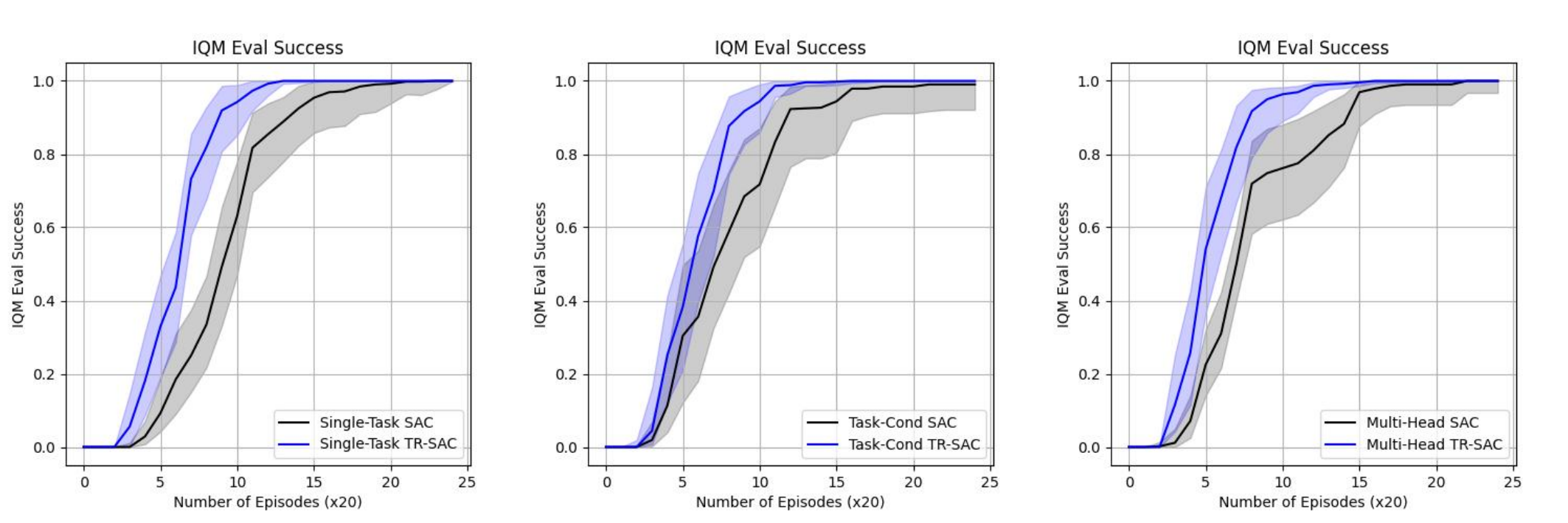}
    \caption{IQM of success rate for multi-task settings in 10 environments from Robosuite with separate plots for each algorithm pair (TR vs. no TR).
    "Task-Cond" and "Multi-Head" are short for "task-conditioned" and "multi-headed" respectively.}
    \label{fig:robosuite10_separate}
\end{figure}

\begin{figure}
    \centering
    \includegraphics[width=0.9\linewidth]{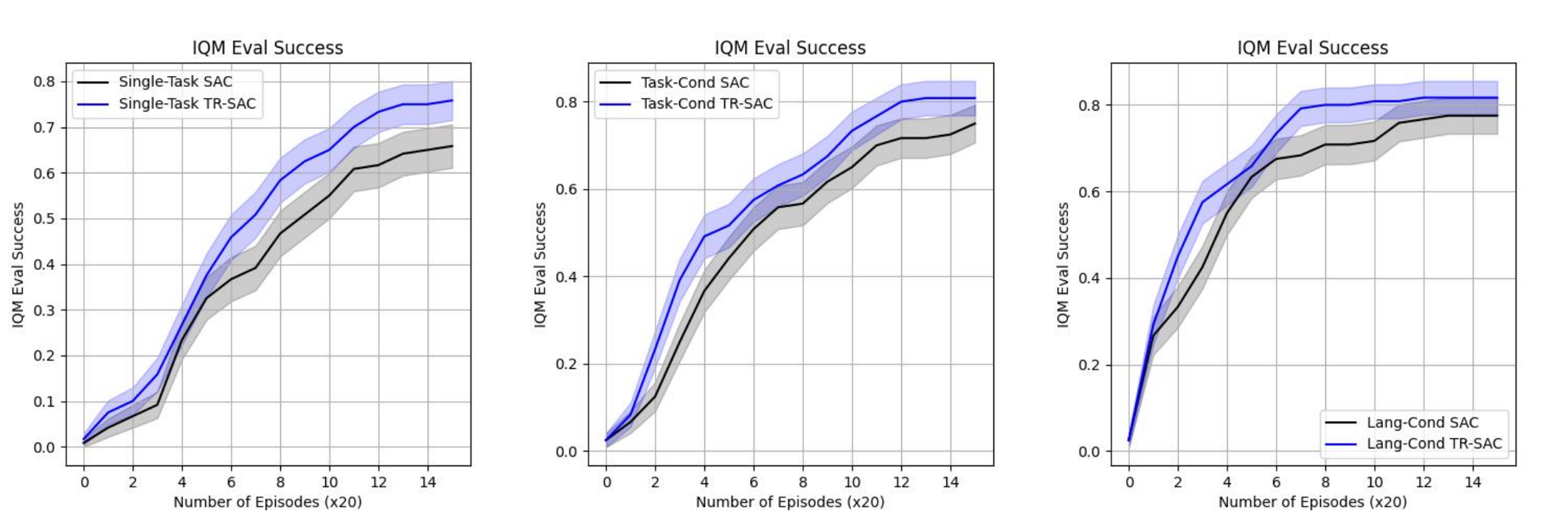}
    \caption{IQM of success rate for multi-task settings in 12 pair of reversible tasks in MT50 of Meta-World with separate plots for each algorithm pair (TR vs. no TR).
    "Task-Cond" and "Lang-Cond" are short for "task-conditioned" and "language-conditioned" respectively.}
    \label{fig:metaworld_reversible24_separate}
\end{figure}

Due to the page limit in the main text, we combine the results of all methods into a single plot for \Cref{fig:IQM_robosuite10} and \Cref{fig:IQM_reversible24}.
As shown in \Cref{fig:robosuite10_separate} and \Cref{fig:metaworld_reversible24_separate}, we include separate plots for each algorithm pair (TR vs. no TR) for better readability.

\section{Ablation Study of Potential Models}
\label{appendix: tune_rev_pot}
\begin{figure}
    \centering
    \includegraphics[width=0.45\linewidth]{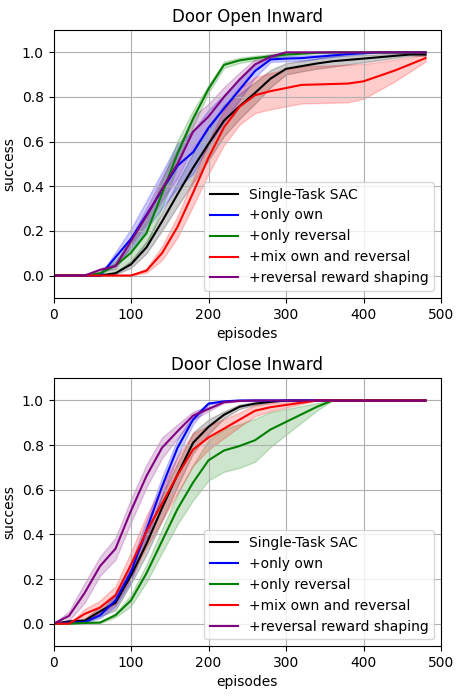}
    \caption{Ablation study of potential models in the task pair of door opening/closing inward.
    "Single-Task SAC" serves as the baseline.
    "+only own" indicates using only the task’s own successful trajectories to train one potential model.
    "only reversal" indicates using only the reversible task’s trajectories to train one potential model.
    "mix own and reversal" indicates training a joint potential model with successful trajectories from two tasks.
    "+reversal reward shaping" indicates training two separate potential models, one for successful trajectories from its own and the other for successful trajectories from the reversible task.}
    \label{fig:ablation pot}
\end{figure}

In our ablation study of potential models, we evaluated four training strategies: (1) using only the task’s own successful trajectories to train one potential model, (2) using only the reversible task’s trajectories to train one potential model, (3) training a joint potential model with successful trajectories from two tasks, and (4) training two separate potential models, one for successful trajectories from its own and the other for successful trajectories from the reversible task. 
The final reward value is then computed as the average of the rewards obtained from these two models. 
Based on the ablation study shown in \Cref{fig:ablation pot}, we conclude that training two potential models is always better than the baseline.
Therefore, we use this setting to train potential models in subsequent experiments.

\section{Ablation Study of Potential Value Labeling Function in Potential-Based Reward Shaping}
\label{appendix: pot_tune}
\begin{figure}
    \centering
    \includegraphics[width=0.99\linewidth]{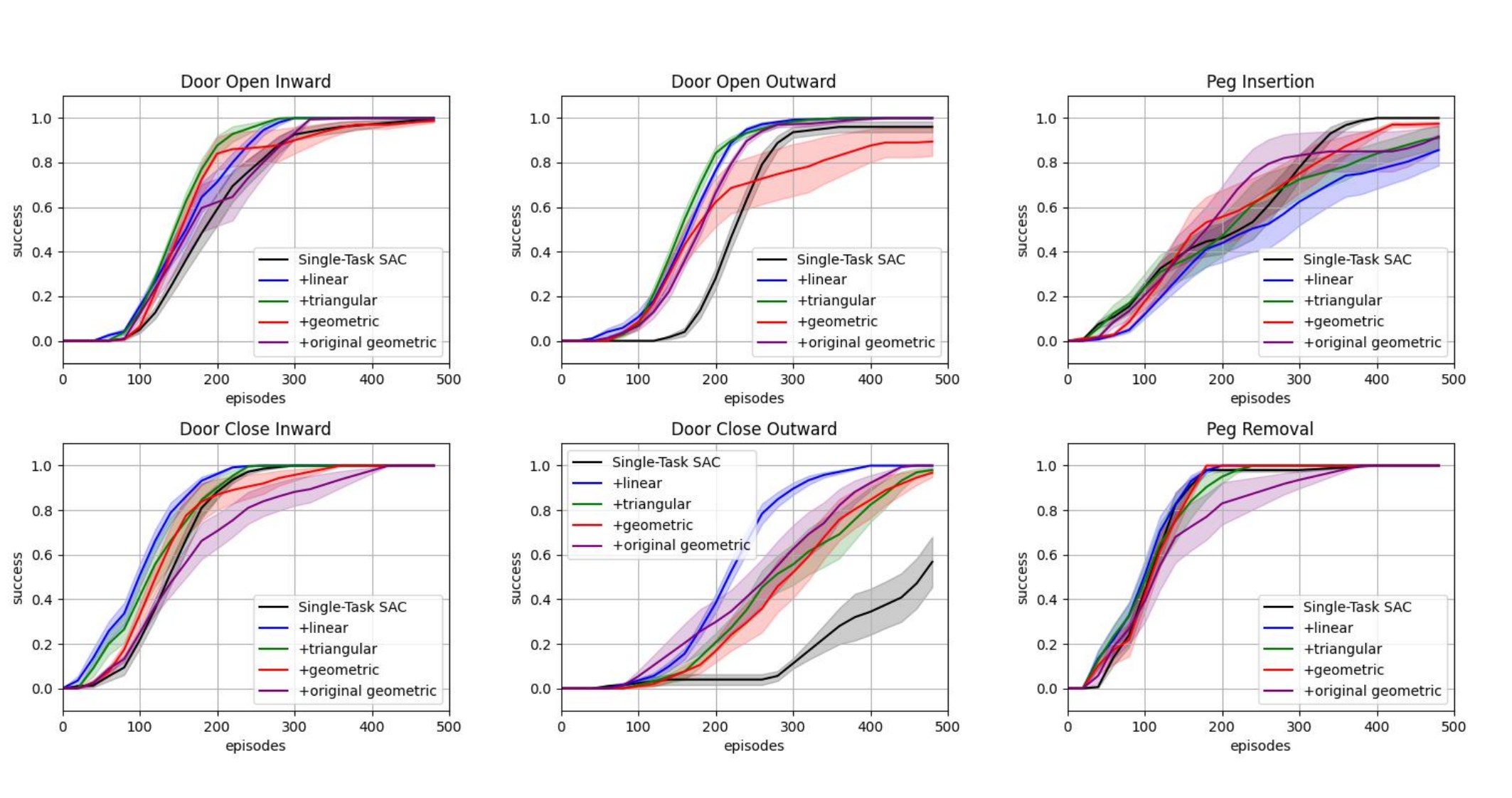}
    \caption{Evaluation curves of agent success rate using time reversal symmetry guided reward shaping with different potential types.}
    \label{fig:rev_pot_10demo_compare_pot_type}
\end{figure}

\begin{figure}
    \centering
    \includegraphics[width=0.35\linewidth]{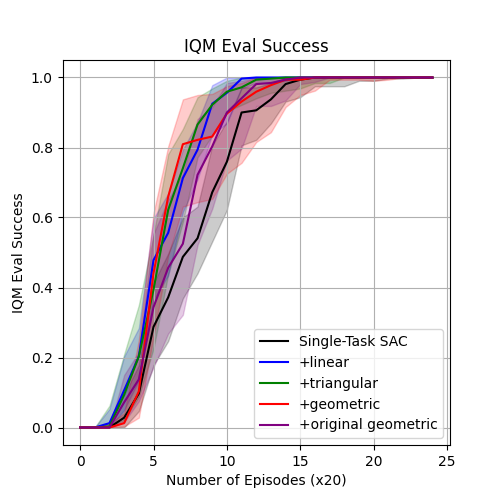}
    \caption{IQM of agent success rate using time reversal symmetry guided reward shaping with different potential types.}
    \label{fig:IQM_rev_pot_10demo_compare_pot_type}
\end{figure}

We evaluate four monotonically increasing functions as the potential value labeling function for a successful trajectory of length $n$.
\begin{itemize}
    \item Linear: ${\Potential}(s_t)=\frac{t}{n}$,
    \item Triangular: ${\Potential}(s_t)=\frac{t(t+1)}{n(n+1)}$,
    \item Original Geometric: ${\Potential}(s_t)=\gamma^{n-t}$,
    \item Geometric: ${\Potential}(s_t)=\frac{\gamma^{n-t}-\gamma^{n-1}}{1-\gamma^{n-1}}$.
\end{itemize}
Results are shown in \Cref{fig:rev_pot_10demo_compare_pot_type}, with inter-quartile mean (IQM) in \Cref{fig:IQM_rev_pot_10demo_compare_pot_type}.
Based on these results, we adopt the linear function for potential-based reward shaping in subsequent experiments.

\section{Results of Time Reversal Symmetry Guided Reward Shaping in Robosuite}
\begin{figure}
    \centering
    \includegraphics[width=0.8\linewidth]{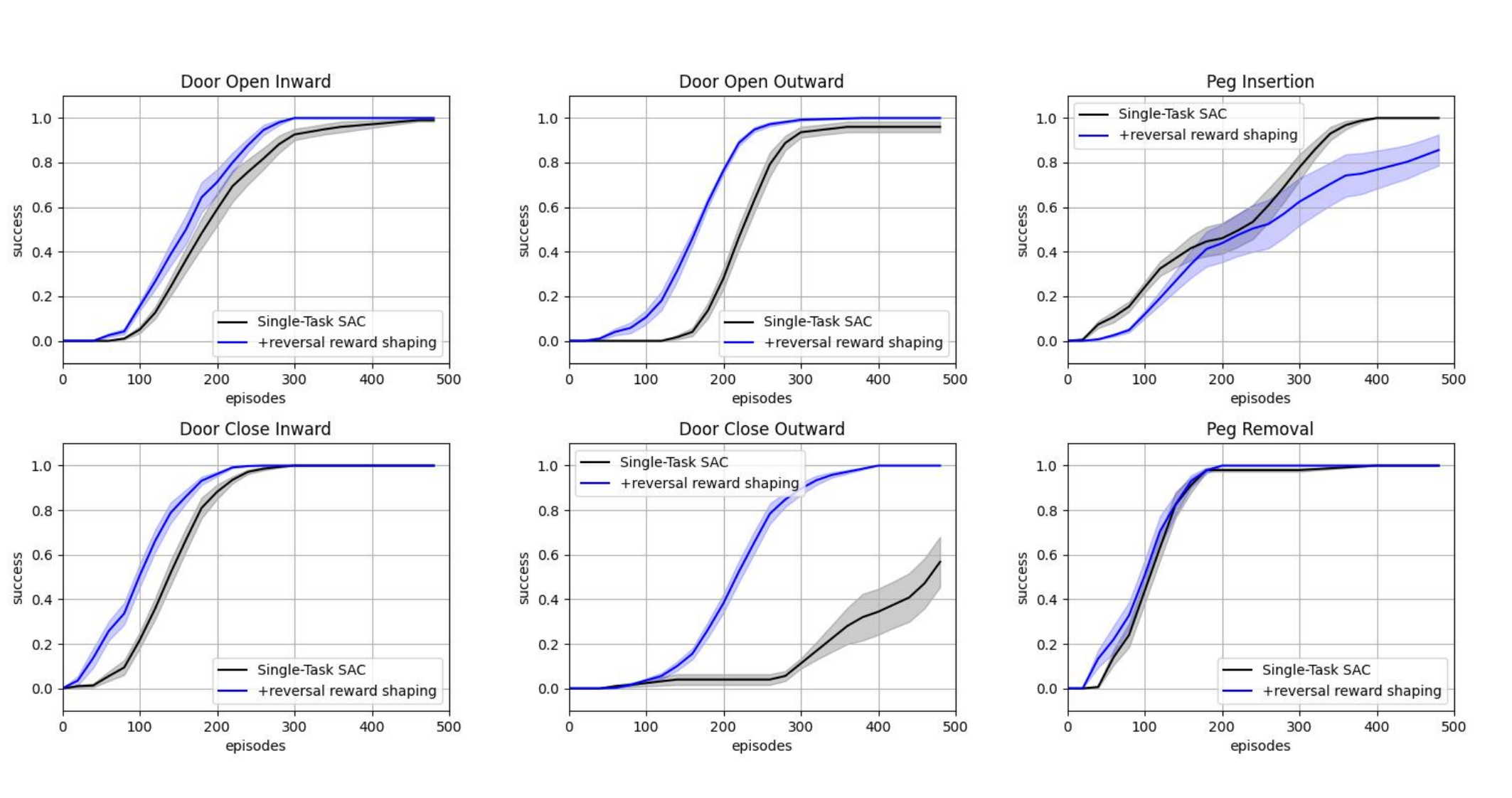}
    \caption{Evaluation curves of time reversal symmetry guided reward shaping in 6 environments of robosuite.
    "Single-Task SAC" serves as the baseline. 
    "+reversal reward shaping" introduces time reversal symmetry guided reward shaping.}
    \label{fig:rev_pot_10demo}
\end{figure}

As shown in \Cref{fig:rev_pot_10demo}, we plot the evaluation curves of time reversal symmetry guided reward shaping in 6 environments of robosuite, from which we can see clear performance gap between the baseline and using time reversal symmetry guided reward shaping.

\section{Results of Both Proposed Techniques in Robosuite}
\begin{figure}
    \centering
    \includegraphics[width=0.99\linewidth]{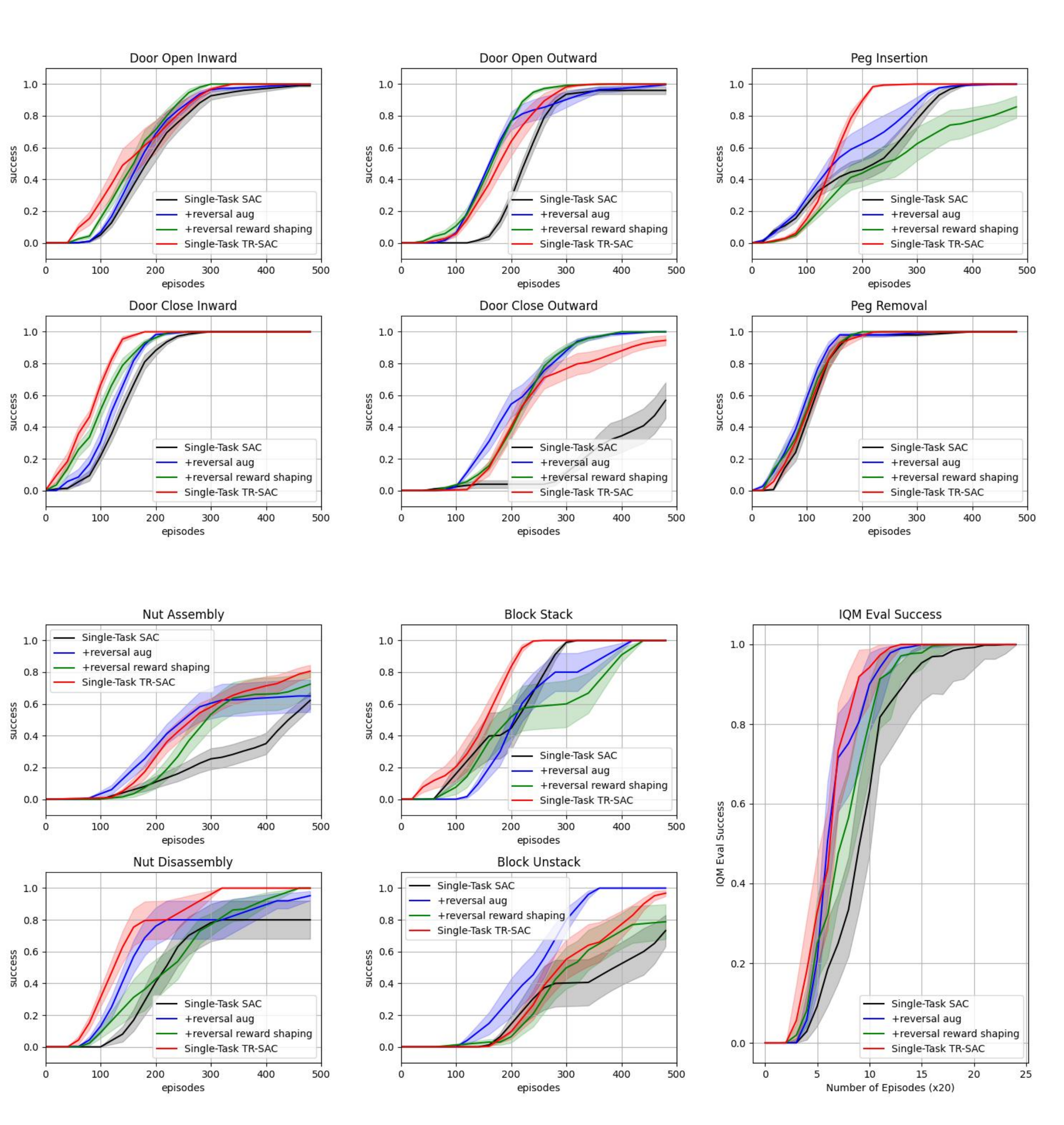}
    \caption{Evaluation curves of both components in 10 environments of Robosuite.
    "reversal aug" represents incorporating reversal augmentation with filtering.
    "reversal reward shaping" represents incorporating potential-based reward shaping.
    "Single-Task TR-SAC" represents our proposed method which combines both components.}
    \label{fig:rev_aug_filter_rev_pot}
\end{figure}

Full evaluation curves of combining trajectory reversal augmentation with dynamics-aware filtering and time reversal symmetry guided reward shaping are provided in \Cref{fig:rev_aug_filter_rev_pot}, which confirm that combining both techniques yields superior performance compared to using either component alone or the baseline method.
\label{appendix: robosuite_rev_aug_filter_rev_pot}

\section{Results of multi-task settings in Robosuite}
\begin{figure}
    \centering
    \includegraphics[width=0.99\linewidth]{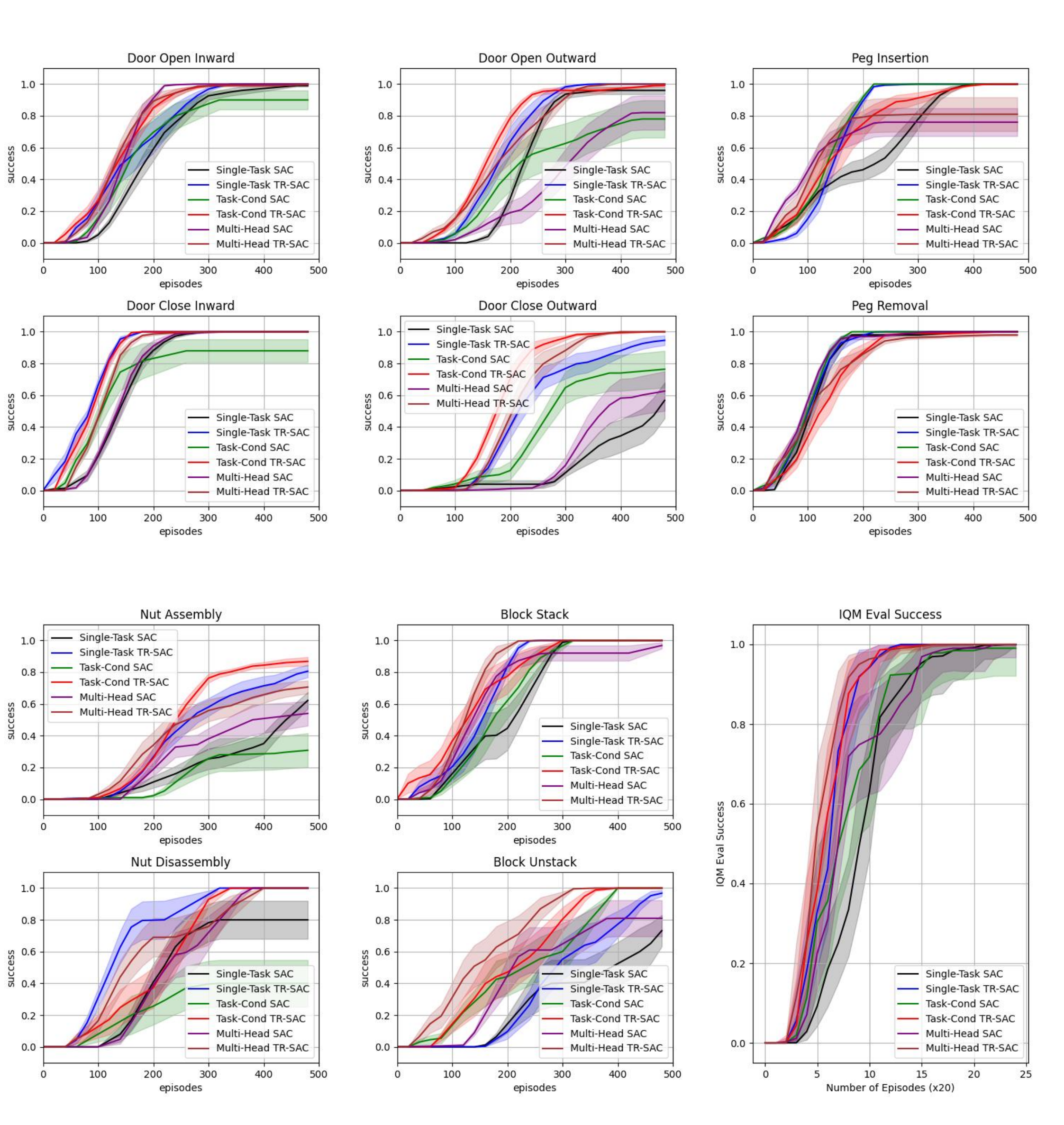}
    \caption{Evaluation curves for multi-task settings in 10 environments of Robosuite.
    "Task-Cond" and "Multi-Head" are short for "task-conditioned" and "multi-headed" respectively.}
    \label{fig:robosuite_mt}
\end{figure}

Full evaluation curves of agent performance in 10 environments of Robosuite are shown in \Cref{fig:robosuite_mt}.
\label{appendix: robosuite_mt}

\section{Additional results of MT50 in Meta-World}
\begin{figure}
\begin{minipage}{0.99\linewidth}
    \centering
    \includegraphics[width=0.99\linewidth]{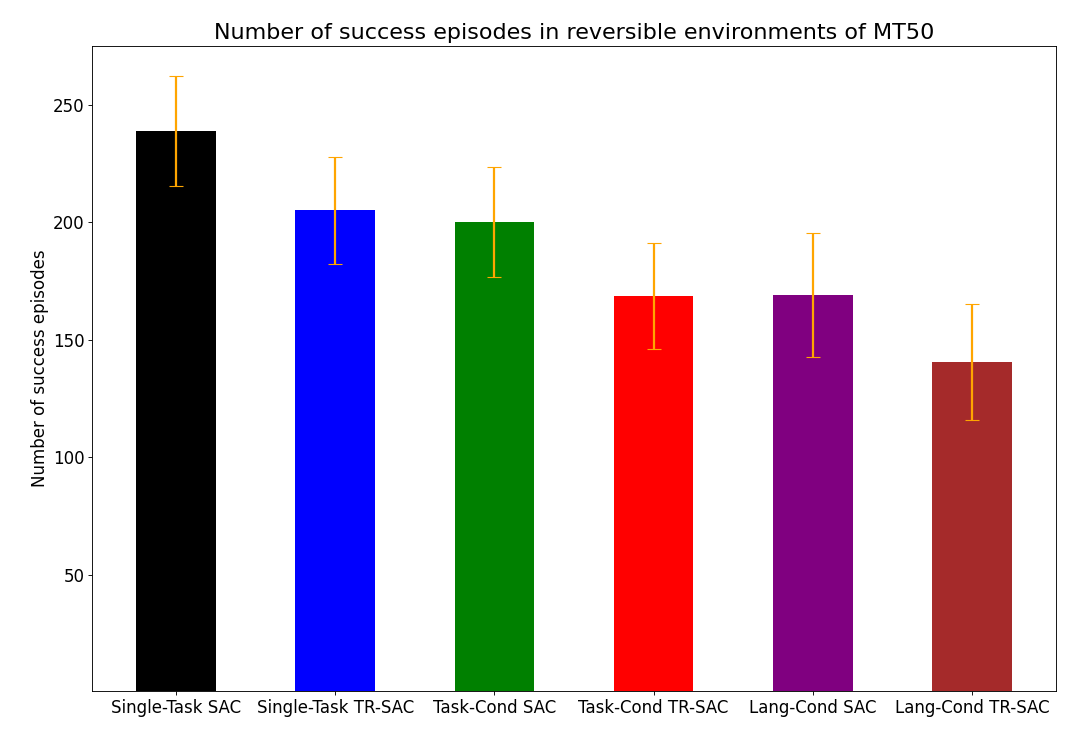}
    \caption{Average number of training episodes required to achieve a $100\%$ success rate in 12 pairs of reversible environments of MT50..}
    \label{fig:num_success_episode_metworld_reversible24}   
\end{minipage}
\begin{minipage}{0.45\linewidth}
    \centering
    \includegraphics[width=0.99\linewidth]{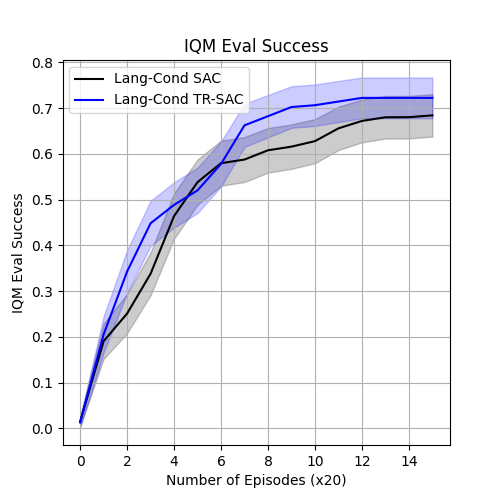}
    \caption{IQM for agent success rate in all 50 environments of MT50.}
    \label{fig:IQM_metaworld_MT50}    
\end{minipage}
\hfill
\begin{minipage}{0.45\linewidth}
    \centering
    \includegraphics[width=0.99\linewidth]{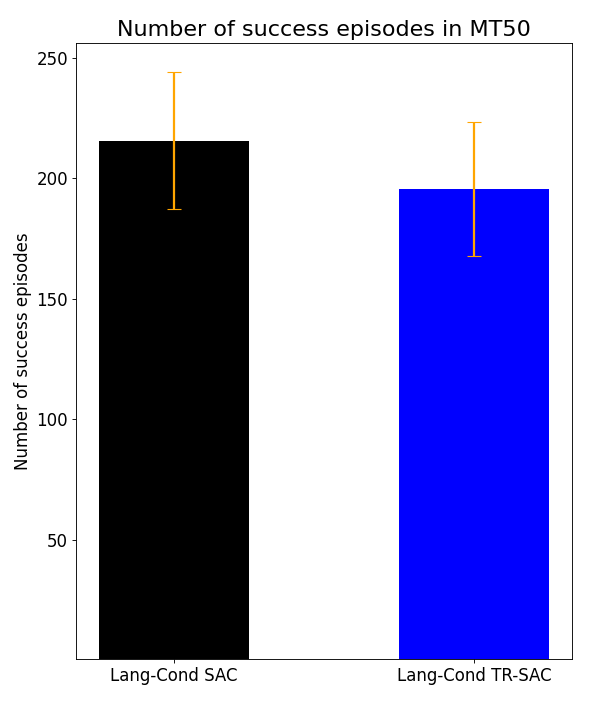}
    \caption{Average number of training episodes required to achieve a $100\%$ success rate in all 50 environments of MT50.}
    \label{fig:num_success_episode_metworld_MT50}
\end{minipage}
\end{figure}

Additional results of agent performance in both 12 reversible task pairs and all 50 environments of MT50 are shown in \Cref{fig:num_success_episode_metworld_reversible24}, \Cref{tab:num_success_episode_metworld_reversible24}, \Cref{fig:IQM_metaworld_MT50}, and \Cref{fig:num_success_episode_metworld_MT50}.

\begin{table}[ht]
    \centering
    \caption{
    Number of training episodes required for $100\%$ success rate for 12 pairs of reversible environments in MT50. 
    Each value is averaged over five runs, with the mean and standard deviation reported.
    "Task-Cond" and "Lang-Cond" are short for "task-conditioned" and "language-conditioned" respectively.
    \textbf{Lower is better.}}
    \small
    \begin{tabular}{c c c c c c c}
        \toprule
        Environment &\makecell{Single-Task\\SAC} & \makecell{Single-Task\\TR-SAC} & \makecell{Task-Cond\\SAC} & \makecell{Task-Cond\\TR-SAC} & \makecell{Lang-Cond\\SAC} & \makecell{Lang-Cond\\TR-SAC}\\
        \midrule
        assembly & \makecell{460$\pm$7} & \makecell{432$\pm$13} & \makecell{456$\pm$8} & \makecell{424$\pm$21} & \makecell{340$\pm$28} & \textbf{\makecell{252$\pm$29}} \\
        \midrule
        disassemble & 500$\pm$0 & 500$\pm$0 & 500$\pm$0 & 500$\pm$0 & 424$\pm$21 & \textbf{364$\pm$24} \\ 
        \midrule
        coffee pull & 444$\pm$16 & \textbf{408$\pm$16} & 460$\pm$4 & 468$\pm$9 & 416$\pm$24 &  416$\pm$24\\ 
        \midrule
        coffee push & 288$\pm$16 & 236$\pm$19 & 316$\pm$24 & 224$\pm$10 & 228$\pm$32 &  \textbf{156$\pm$25} \\ 
        \midrule
        door lock & 360$\pm$26 & 352$\pm$23 & 112$\pm$13 & 88$\pm$5 & 72$\pm$13 & \textbf{60$\pm$7} \\ 
        \midrule
        door unlock & 196$\pm$22 & 148$\pm$14 & 144$\pm$7 & 88$\pm$9 & 96$\pm$10 & \textbf{72$\pm$5} \\ 
        \midrule
        door open & 252$\pm$11 & 208$\pm$16 & 208$\pm$11 & \textbf{168$\pm$7} & 204$\pm$22 & 252$\pm$29 \\ 
        \midrule
        door close & 100$\pm$4 & 88$\pm$1 & 80$\pm$2 & 48$\pm$1 & 52$\pm$4 & \textbf{44$\pm$5} \\ 
        \midrule
        drawer open & 284$\pm$13 & 200$\pm$9 & 272$\pm$26 & 172$\pm$15 & 140$\pm$8 & \textbf{104$\pm$12} \\ 
        \midrule
        drawer close & 40$\pm$3 & 16$\pm$1 & 16$\pm$2 & 12$\pm$1 & \textbf{8$\pm$1} & \textbf{8$\pm$1} \\ 
        \midrule
        faucet open & 96$\pm$2 & 84$\pm$1 & 92$\pm$5 & 80$\pm$3 & 48$\pm$5 & \textbf{40$\pm$6} \\ 
        \midrule
        faucet close & 124$\pm$6 & 156$\pm$7 & 72$\pm$2 & \textbf{60$\pm$3} & 76$\pm$2 & 64$\pm$5 \\ 
        \midrule
        handle press & 32$\pm$4 & 20$\pm$2 & 20$\pm$2 & 24$\pm$3 & 24$\pm$1 & \textbf{20$\pm$0} \\ 
        \midrule
        handle pull & 244$\pm$14 & 172$\pm$9 & 308$\pm$20 & \textbf{128$\pm$7} & 168$\pm$24 & 160$\pm$25 \\ 
        \midrule
        \makecell{peg insert\\ side} & 328$\pm$22 & 296$\pm$18 & 348$\pm$18 & \textbf{276$\pm$16} & 424$\pm$21 & 348$\pm$26 \\ 
        \midrule
        \makecell{peg unplug\\ side} & 164$\pm$13 & 128$\pm$9 & 76$\pm$3 & 68$\pm$3 & 180$\pm$24 & \textbf{32$\pm$1} \\ 
        \midrule
        plate slide & 208$\pm$14 & 180$\pm$14 & 176$\pm$4 & 104$\pm$11 & 80$\pm$4 & \textbf{64$\pm$7} \\ 
        \midrule
        \makecell{plate slide\\ back} & 256$\pm$18 & 224$\pm$20 & 152$\pm$10 & 84$\pm$3 & 40$\pm$4 & \textbf{36$\pm$2} \\ 
        \midrule
        \makecell{plate slide\\ side} & 228$\pm$16 & 156$\pm$13 & 128$\pm$9 & 148$\pm$20 & 80$\pm$7 & \textbf{60$\pm$6} \\ 
        \midrule
        \makecell{plate slide\\ back side} & 196$\pm$12 & 148$\pm$4 & 140$\pm$7 & 152$\pm$8 & 104$\pm$9 & \textbf{72$\pm$5} \\ 
        \midrule
        push & 288$\pm$17 & \textbf{140$\pm$10} & 272$\pm$24 & 244$\pm$18 & 404$\pm$27 & 328$\pm$30 \\ 
        \midrule
        push back & 480$\pm$6 & 500$\pm$0 & \textbf{316$\pm$22} & 396$\pm$14 & 368$\pm$24 & 324$\pm$30 \\ 
        \midrule
        window open & 84$\pm$1 & 64$\pm$1 & 76$\pm$5 & 52$\pm$2 & \textbf{24$\pm$1} & 44$\pm$5 \\ 
        \midrule
        window close & 80$\pm$2 & 68$\pm$1 & 64$\pm$1 & \textbf{40$\pm$0} & 56$\pm$5 & 52$\pm$2 \\ 
        \midrule
        \textbf{ALL} & 239$\pm$23 & 205$\pm$23 & 200$\pm$24 & 169$\pm$23 & 169$\pm$26 & \textbf{140$\pm$25} \\
        \bottomrule
    \end{tabular}
    \label{tab:num_success_episode_metworld_reversible24}
\end{table}

\begin{table}[ht]
    \centering
    \caption{
    Number of training episodes required for $100\%$ success rate for all 50 environments in MT50. 
    Each value is averaged over five runs, with the mean and standard deviation reported.
    "Lang-Cond" is short for "language-conditioned".
    \textbf{Lower is better.}}
    \small
    \begin{tabular}{c c c | c c c}
        \toprule
        Environment & \makecell{Lang-Cond\\SAC} & \makecell{Lang-Cond\\TR-SAC} & Environment & \makecell{Lang-Cond\\SAC} & \makecell{Lang-Cond\\TR-SAC}\\
        \midrule
        assembly & \makecell{340$\pm$28} & \textbf{\makecell{252$\pm$29}} & sweep-into & 276$\pm$27 & \textbf{172$\pm$24} \\
        \midrule
        disassemble & 424$\pm$21 & \textbf{364$\pm$24} & reach & \textbf{104$\pm$4} & 164$\pm$24 \\ 
        \midrule
        coffee pull & 416$\pm$24 &  416$\pm$24 & reach wall & \textbf{128$\pm$11} & 188$\pm$22\\ 
        \midrule
        coffee push & 228$\pm$32 &  \textbf{156$\pm$25} & stick-pull & 172$\pm$23 & \textbf{160$\pm$25} \\ 
        \midrule
        door lock & 72$\pm$13 & \textbf{60$\pm$7} & sweep & 316$\pm$23 & \textbf{284$\pm$25} \\ 
        \midrule
        door unlock & 96$\pm$10 & \textbf{72$\pm$5} & basketball & 500$\pm$0 & 500$\pm$0 \\ 
        \midrule
        door open & \textbf{204$\pm$22} & 252$\pm$29 & bin picking & \textbf{348$\pm$26} & 360$\pm$24 \\ 
        \midrule
        door close & 52$\pm$4 & \textbf{44$\pm$5} & box close & 352$\pm$27 & \textbf{296$\pm$25} \\ 
        \midrule
        drawer open & 140$\pm$8 & \textbf{104$\pm$12} & coffee button & \textbf{52$\pm$3} & 64$\pm$7 \\ 
        \midrule
        drawer close & \textbf{8$\pm$1} & \textbf{8$\pm$1} & button press & \textbf{28$\pm$2} & 36$\pm$2 \\ 
        \midrule
        faucet open & 48$\pm$5 & \textbf{40$\pm$6} & button press wall & \textbf{76$\pm$2} & 76$\pm$6 \\ 
        \midrule
        faucet close & 76$\pm$2 & \textbf{64$\pm$5} & button press topdown & 220$\pm$32 & \textbf{100$\pm$8} \\ 
        \midrule
        handle press & 24$\pm$1 & \textbf{20$\pm$0} & \makecell{button press\\topdown wall} & 160$\pm$25 & \textbf{100$\pm$9} \\ 
        \midrule
        handle pull & 168$\pm$24 & \textbf{160$\pm$25} & dial turn & 384$\pm$21 & \textbf{356$\pm$25} \\ 
        \midrule
        handle pull side & \textbf{304$\pm$24} & 320$\pm$31 & handle press side & \textbf{32$\pm$3} & 36$\pm$2 \\ 
        \midrule
        peg insert side & 424$\pm$21 & \textbf{348$\pm$26} & hammer & 172$\pm$23 & \textbf{160$\pm$25} \\ 
        \midrule
        peg unplug side & 180$\pm$24 & \textbf{32$\pm$1} & hand insert & \textbf{236$\pm$31} & 256$\pm$28 \\ 
        \midrule
        plate slide & 80$\pm$4 & \textbf{64$\pm$7} & lever pull & 280$\pm$27 & \textbf{252$\pm$30} \\ 
        \midrule
        plate slide back & 40$\pm$4 & \textbf{36$\pm$2} & pick out of hole & \textbf{424$\pm$21} & 428$\pm$20 \\ 
        \midrule
        plate slide side & 80$\pm$7 & \textbf{60$\pm$6} & pick place & 500$\pm$0 & 500$\pm$0 \\ 
        \midrule
        plate slide back side & 104$\pm$9 & \textbf{72$\pm$5} & pick place wall & 416$\pm$24 & 416$\pm$24 \\ 
        \midrule
        push & 404$\pm$27 & \textbf{328$\pm$30} & push wall & \textbf{344$\pm$28} & 356$\pm$25 \\ 
        \midrule
        push back & 368$\pm$24 & \textbf{324$\pm$30} & shelf place & 428$\pm$20 & 428$\pm$20 \\ 
        \midrule
        window open & \textbf{24$\pm$1} & 44$\pm$5 & soccer & 336$\pm$29 & \textbf{248$\pm$29} \\ 
        \midrule
        window close & 56$\pm$5 & \textbf{52$\pm$2} & stick push & \textbf{140$\pm$9} & 156$\pm$6 \\ 
        \midrule
         & \multicolumn{2}{c}{Lang-Cond SAC} & \multicolumn{2}{c}{\makecell{Lang-Cond TR-SAC}} \\
        \midrule
        \textbf{ALL} & \multicolumn{2}{c}{216$\pm$28} & \multicolumn{2}{c}{\textbf{196$\pm$28}} \\
        \bottomrule
    \end{tabular}
    \label{tab:num_success_episode_metworld_MT50}
\end{table}


\section{Compute Resources We Use}
\label{appendix:compute}
In all our experiments, we utilize a GPU server equipped with 8 cards that have either RTX-4090 or A6000 GPUs and are powered by AMD EPYC 7763 CPUs.
For experiments in robosuite: training a single-task agent takes around 5 hours while training a multi-task agent for two tasks takes around 10 hours for 500 training episodes for each task.
For experiments in MetaWorld: training a single-task agent takes around 2 hours while training a multi-task agent for two tasks takes around 4 hours for 500 training episodes for each task.
For MT50, it takes around 3 days to train an agent that handles 50 tasks.

\section{Limitations}
\label{appendix: limitations}
A key limitation of our work is the absence of real-robot experiments, as our current experiments are all conducted in simulation environments. 
While simulations enable efficient prototyping and scalability, they may oversimplify physical dynamics, sensor noise, or actuator constraints inherent in real-world robotic systems. 
Future work could address this gap by deploying the proposed method on physical robots, ensuring robustness and generalizability to practical applications. 
\jypmodify{Like other data augmentation methods, our approach relies on prior knowledge of task structures—specifically, time reversal symmetry in our case.}

\section{Broader Impacts}
\label{appendix: broader impacts}
\paragraph{Positive societal impacts}: 
This work advances the sample efficiency of deep reinforcement learning (DRL) agents in robotics manipulation tasks, enabling faster and more cost-effective training for practical applications. 
By reducing the computational resources required for training, it lowers barriers to deploying robotic systems in real-world setting.
Improved sample efficiency also minimizes energy consumption and hardware wear, aligning with sustainability goals.
Furthermore, robust and efficient DRL methodologies can accelerate the development of autonomous systems that enhance productivity, safety, and accessibility, ultimately contributing to economic growth and societal well-being.

\paragraph{Negative societal impacts}
To the best of our knowledge, we don't see any negative societal impacts of our work.


\end{document}